\documentclass[10pt,twocolumn,letterpaper]{article}

\usepackage{cvpr}
\usepackage{times}
\usepackage{epsfig}
\usepackage{graphicx}
\usepackage{amsmath}
\usepackage{amssymb}
\usepackage{lipsum}
\usepackage{import}
\usepackage{cite}
 \usepackage{multirow}
 \usepackage{booktabs}
\usepackage{textcomp}
\usepackage[dvipsnames]{xcolor}

\usepackage[pagebackref=true,breaklinks=true,letterpaper=true,colorlinks,bookmarks=false]{hyperref}

\cvprfinalcopy %

\def\cvprPaperID{2164} %

\newcommand{\ag}[1]{\textcolor{ForestGreen}{#1}}
\newcommand{\vb}[1]{\textcolor{OrangeRed}{#1}}
\newcommand{\pat}[1]{\textcolor{RoyalBlue}{#1}}
\newcommand{\inst}[1]{\textcolor{Magenta}{#1}}
\newcommand{\loc}[1]{\textcolor{Sepia}{#1}}
\newcommand{\atmp}[1]{\textcolor{violet}{#1}}
\newcommand{\adir}[1]{{#1}}

\newcommand{\dt}[1]{\texttt{#1}}
\newcommand{\arch}[0]{VOGNet\xspace}
\newcommand{\tk}[0]{VOG\xspace}

\newcommand{\sep}[0]{\texttt{SEP}\xspace}
\newcommand{\temp}[0]{\texttt{TEMP}\xspace}
\newcommand{\spat}[0]{\texttt{SPAT}\xspace}

\newcommand{\R}{\mathbb{R}}

\newcommand{\acs}[0]{ActivityNet-SRL\xspace}

\newcommand{\bsi}[0]{ImgGrnd\xspace}
\newcommand{\bsv}[0]{VidGrnd\xspace}

\newcommand{\gt}[0]{GT5\xspace}
\newcommand{\phun}[0]{P100\xspace}

\newcommand{\svsq}[0]{\texttt{SVSQ}\xspace}

\newcommand{\acc}[0]{Acc\xspace}
\newcommand{\vacc}[0]{VAcc\xspace}
\newcommand{\sacc}[0]{SAcc\xspace}
\newcommand{\cons}[0]{Cons\xspace}

\newcommand{\tpr}[0]{\toprule\toprule}
\newcommand{\btr}[0]{\bottomrule\bottomrule}
\belowdisplayskip \abovedisplayskip

\newlength{\abstractReduceTop}
\newlength{\abstractReduceBot}

\newlength{\sectionReduceTop}
\newlength{\sectionReduceBot}

\newlength{\subsectionReduceTop}
\newlength{\subsectionReduceBot}

\newlength{\subsubsectionReduceTop}
\newlength{\subsubsectionReduceBot}

\newlength{\captionReduceTop}
\newlength{\captionReduceBot}

\newlength{\eqnReduceTop}
\newlength{\eqnReduceBot}

\newlength{\horSkip}
\newlength{\verSkip}

\newlength{\figureHeight}
\begin{document}

\title{Video Object Grounding using Semantic Roles in Language Description}

\author{Arka Sadhu$^1$ \quad \quad Kan Chen$^{2}$ \quad \quad Ram Nevatia$^1$\\
$^1$University of Southern California \quad \quad $^2$Facebook Inc.\\
{\tt\small {\{asadhu|nevatia\}@usc.edu} \quad kanchen18@fb.com} 
}

\maketitle

\vspace{\abstractReduceTop}
\begin{abstract}
\vspace{\abstractReduceBot}

We explore the task of Video Object Grounding (\tk), which grounds objects in videos referred to in natural language descriptions. 
Previous methods apply image grounding based algorithms to address \tk, fail to explore the object relation information and suffer from limited generalization. 
Here, we investigate the role of object relations in \tk and propose a novel framework \arch to encode multi-modal object relations via self-attention with relative position encoding.
To evaluate \arch, we propose novel contrasting sampling methods to generate more challenging grounding input samples, and construct a new dataset called ActivityNet-SRL (ASRL) based on existing caption and grounding datasets. 
Experiments on ASRL validate the need of encoding object relations in \tk, and our \arch outperforms competitive baselines by a significant margin.

\end{abstract}
\vspace{\abstractReduceBot}

\vspace{\sectionReduceTop}
\section{Introduction}
\vspace{\sectionReduceTop}
\label{sec:intro}

Grounding objects in images \cite{yu2016modeling,chen2017query,yu2018mattnet} and videos \cite{Khoreva2018VideoOS,Zhou2018WeaklySupervisedVO,Chen2019WeaklySupervisedSG} from natural language queries is a fundamental task at the intersection of Vision and Language.
It is a building block for downstream grounded vision+language tasks such as Grounded-VQA \cite{zhu2016visual7w,Zhang2016AutomaticGO,Lei2018TVQALC,Lei2019TVQASG,Gao2018MotionAppearanceCN}, Grounded-Captioning \cite{Luo2017ComprehensionGuidedRE,Lu2018NeuralBT,Ma2017GroundedOA,zhou2019grounded} and Grounded Navigation \cite{Hu2019AreYL}.

\begin{figure}
    \centering

    \includegraphics[width=\linewidth]{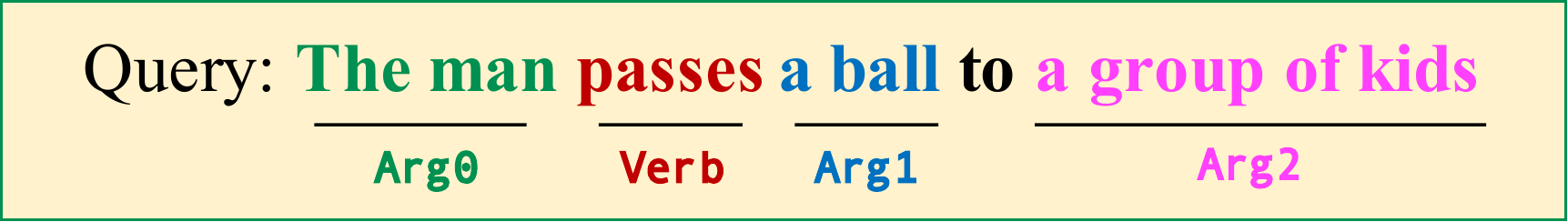}
    \includegraphics[width=\linewidth]{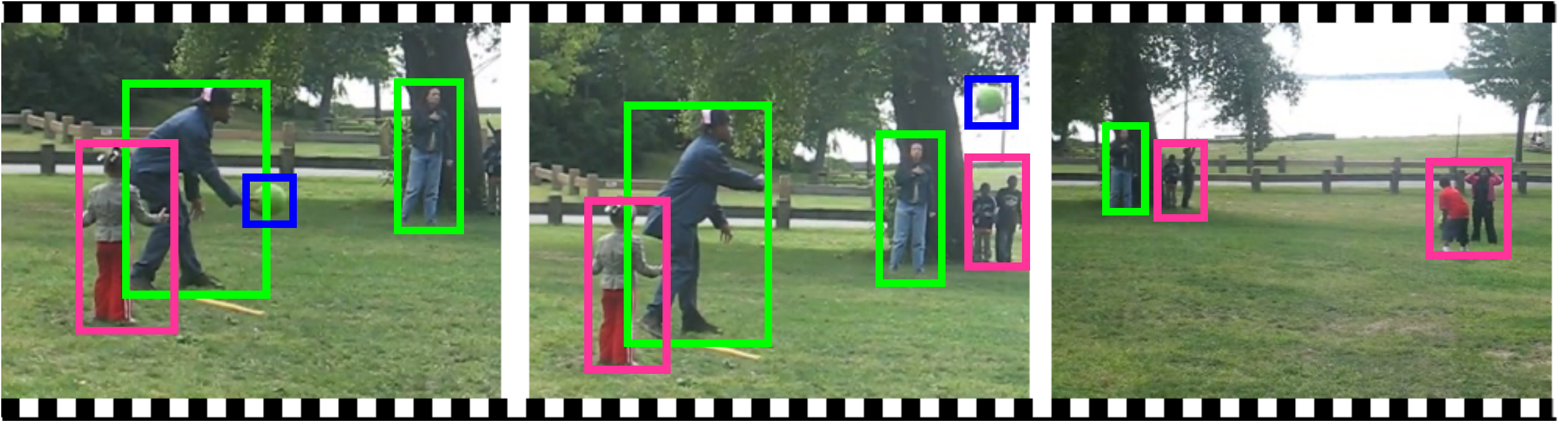}
    \footnotesize{(a) localize individual queries: ``man'', ``ball'', ``kids''} 
    \includegraphics[width=\linewidth]{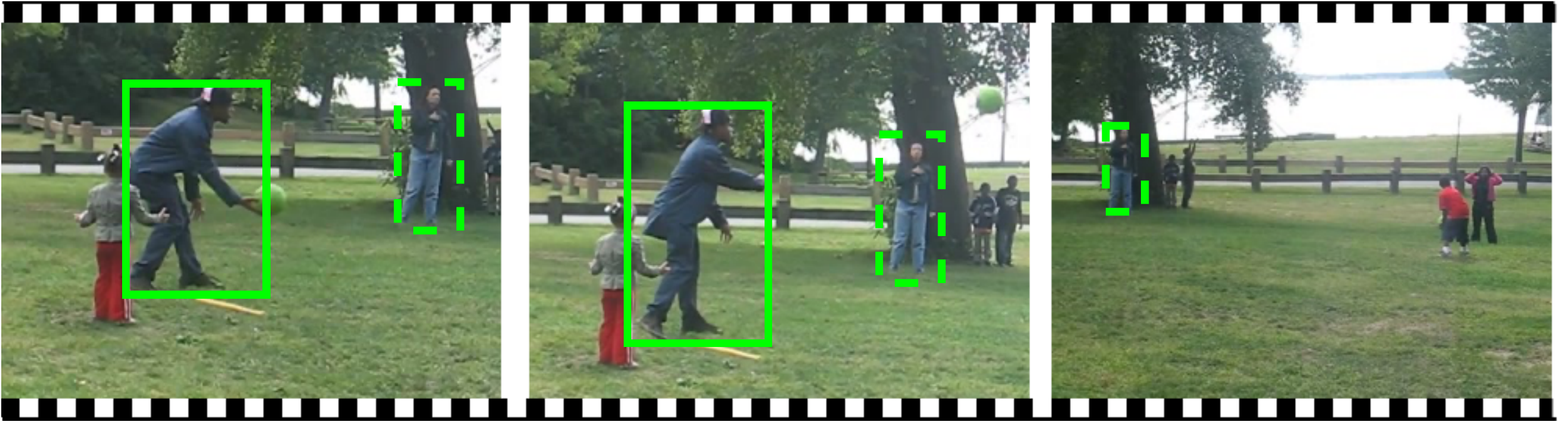}
    \footnotesize{(b) localize the spatio-temporal tube from query uniquely identifying it (``man passing the ball'')}
    \includegraphics[width=\linewidth]{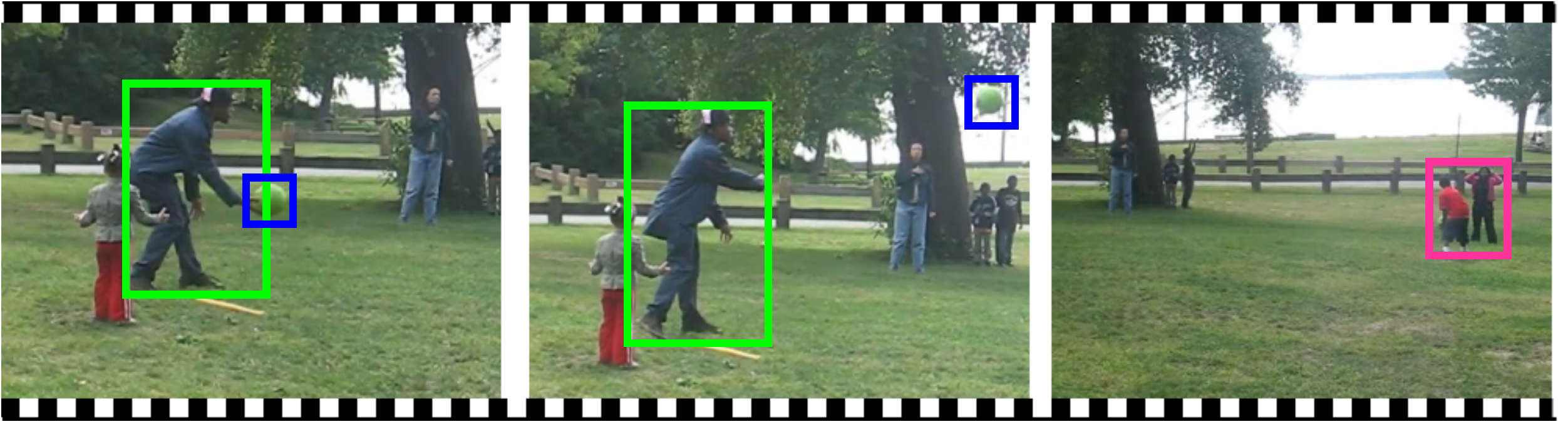}
    \footnotesize{(c) localize only the referred objects in the query \\ (``man'', ``ball'', ``group of kids'')}
    \caption{
    Illustration of different formulations of \tk when the same query and video frames are used.
    (a) \cite{Zhou2018WeaklySupervisedVO} treats each query word independently and doesn't distinguish between different instances of the same object.
    (b) \cite{Chen2019WeaklySupervisedSG} makes this distinction using independent spatio-temporal tubes.
    Ours (c) involves localizing only those objects which are being referenced in the query and requires additional disambiguation using object relations.
    }
    \label{fig:intro}
    \vspace{-3mm}
\end{figure}

In this work, we address the task of Video Object Grounding (\tk):
given a video and its natural language description we aim to localize each referred object.
Different from prior \tk methods on finding objects from query mentions \cite{Zhou2018WeaklySupervisedVO} or distinguishing spatio-temporal tubes from a referring expression \cite{Chen2019WeaklySupervisedSG},
we formulate \tk as localizing only the specific referred objects in the query.
Prior work has focused on attending to each object in isolation; our formulation additionally requires incorporating object-object relations in both time and space.
Figure \ref{fig:intro} illustrates the key differences.

Despite the importance of associating natural language descriptions with objects in videos, \tk has remained relatively unexplored due to two practical requirements:
(i) a large-scale video dataset with object-level annotations,
(ii) the videos should contain multiple instances of the same object category so making a distinction among them becomes necessary. 
Recently, \cite{zhou2019grounded} released ActivityNet-Entities dataset which contains bounding box annotations relating the noun-phrases of the video descriptions \cite{krishna2017dense} to the corresponding objects instances in ActivityNet \cite{caba2015activitynet} videos.
Despite its scale, a majority of the videos in ActivityNet 
contain single instances of various objects.
For instance, in Figure \ref{fig:intro} ``ball'' can be localized simply using an object detection system such as FasterRCNN \cite{ren2015faster} without relating ``ball'' to the ``man'' or the ``kids''.

We mitigate this absence of multiple object instances in two steps.
First, we sample \textit{contrasting examples} from the dataset; these are examples that are similar to but not exactly the same as described by the language query.
To sample contrasting examples, we obtain semantic-roles (SRLs) using a state-of-the-art Semantic Role Labeling (SRL) system \cite{shi2019simple} 
on the language descriptions.
SRLs answer the high-level question of ``who (\ag{\dt{Arg0}}) did what (\vb{\dt{Verb}}) to whom (\pat{\dt{Arg1}})'' \cite{Strubell2018LinguisticallyInformedSF}.
We sample videos with descriptions of the same semantic roles structure as the queried description, but the role is realized by a different noun or a verb.

In the next step, we need to present the contrasting videos to a model.
If the contrasting samples are processed independently, a model could easily ``cheat'' and find the associated video by simply adding the object detection and action recognition scores as per the query. 
To prevent this, we propose novel spatial and temporal concatenation methods to merge contrasting samples into one video.
With contrasting objects and their relations in the same video, the model is forced to encode object relations in order to ground the referred objects (details in Section \ref{ss:bias_strats}).

Clearly, encoding object relations is of primary importance for \tk.
Recently, \cite{girdhar2019video} and \cite{zhou2019grounded} show promising results using self-attention \cite{vaswani2017attention} to encode object relations.
However, there are two issues in directly adapting self-attention on objects for \tk.
First, such object relations are computed independent of the language creating ambiguities when two objects have multiple relations.
For instance, in Figure \ref{fig:intro} ``The man is playing with a group of kids'' is an accurate description for the same video but the queried relation between ``the man'' and ``kids'' is different.
Second, the transformer module for self-attention \cite{vaswani2017attention} expects positional encoding for its input but 
absolute positions are not meaningful in a video.

We address these issues in our proposed  \textbf{\arch} framework which applies self-attention to both the object features and fused multi-modal features to encode language dependent and independent object relations. 
To encode positions, we propose a relative position encoding (RPE) scheme based on \cite{Shaw2018SelfAttentionWR}.
Essentially, RPE biases the model to weigh related objects based on their proximity (details on model architecture in Section \ref{ss:framework}).

To evaluate our models, we contribute \acs which adds semantic roles to the descriptions \cite{krishna2017dense} and aligns with the noun-phrase annotations in \cite{zhou2019grounded}.
We further show by pre-computing lemmatized noun-phrases, contrastive sampling process can be used in training (details on dataset construction in Section~\ref{ss:data_const},\ref{ss:ecs}).

Our contributions are three-fold:
(i) we explore \textbf{\tk} and propose contrastive sampling with temporal and spatial concatenation to allow learning object relations
(ii) we design 
\arch which extends self-attention to encode language-dependent object relations and relative position encodings
(iii) we contribute ActivityNet-SRL as a benchmark for \tk.
Our code and dataset are publicly available\footnote{\url{https://github.com/TheShadow29/vognet-pytorch}}.

\vspace{\sectionReduceTop}
\section{Related Work}
\vspace{\sectionReduceBot}
\label{sec:related}
\textbf{Grounding objects} in images is a heavily studied topic under referring expression \cite{yu2016modeling,yu2018mattnet,mao2016generation,kazemzadeh2014referitgame} and phrase localization \cite{chen2017query,rohrbach2016grounding,plummer2015flickr30k,plummer2018conditional,Sadhu_2019_ICCV}.
In contrast, grounding objects in videos has garnered less interest. 
Apart from \cite{Zhou2018WeaklySupervisedVO,Chen2019WeaklySupervisedSG}, \cite{Khoreva2018VideoOS} enforces temporal consistency for video object segmentation and requires the target to be in each frame and \cite{Huang2018FindingW} use structured representations in videos and language for co-reference resolution.
Different from them, our proposed formulation of \tk elevates the role of object relations and supports supervised training due to use of a larger dataset.

\textbf{Object relations} is also fairly well-studied in images under scene-graph generation \cite{Yang2018GraphRF,li2017scene,krishna2017visual,Newell2017PixelsTG} and human-object interaction \cite{Chao2015HICOAB,Chao2017LearningTD,ronchi2015describing,gao2018ican,gkioxari2018detecting,zhuang2018hcvrd} and referring relations \cite{krishna2018referring}.
However, a majority of the relations are spatial (``to-the-left-of'', ``holding'') with considerable biases caused due to co-occurrence \cite{Zellers2017NeuralMS}.
On the video side, it has been explored for spatio-temporal detection \cite{sun2018actor,baradel2018object,girdhar2019video}.
In particular, \cite{girdhar2019video} showed self-attention using transformers \cite{vaswani2017attention} to be more effective than relation-networks \cite{santoro2017simple} based detectors \cite{sun2018actor}.
For \tk, relation networks would not be effective due to high memory requirements
and 
thus we only explore self-attention mechanism. 
Different from \cite{girdhar2019video}, we use bottom-up features \cite{anderson2018bottom} which don't maintain any order. 
As an alternative, we employ relative position encoding.

\textbf{Video relation detection} \cite{Shang2017VideoVR,Shang2017VideoVR,Tsai2019VideoRR} is closely related to \tk where relations between two objects need to detected across video frames. 
However, the metrics used (recall@50/100) are difficult to interpret.
Moreover, densely annotating the relations is expensive and results in less diverse relations.
In contrast, ours uses sparsely annotated frames and leverages off-the-shelf SRL systems.

\textbf{Visual Semantic Role Labeling} in images has focused on situation recognition \cite{yatskar2016situation,yatskar2017commonly,silberer2018grounding}.
To annotate the images, \cite{yatskar2016situation} employed FrameNet\cite{fillmore2003background} annotations
and \cite{silberer2018grounding} shows using semantic parsers on image captions significantly reduces annotation cost.
We instead PropBank annotations \cite{palmer2005proposition} which is verb-oriented and thus more suited to video descriptions.
Finally, our use of semantic roles is guided by contrastive sampling and not assigning semantic roles to visual entities.

\textbf{Contrastive Training} via max-margin loss has been commonly used in vision+language tasks \cite{yu2018mattnet,karpathy2015deep,Zhou2018WeaklySupervisedVO,zhang2019graphical}.
Here, we don't use contrastive losses, instead, the concatenation of the videos directly informs us which objects are related.
As such, we train using binary cross-entropy.

\begin{figure*}
    \centering
    \includegraphics[width=\linewidth]{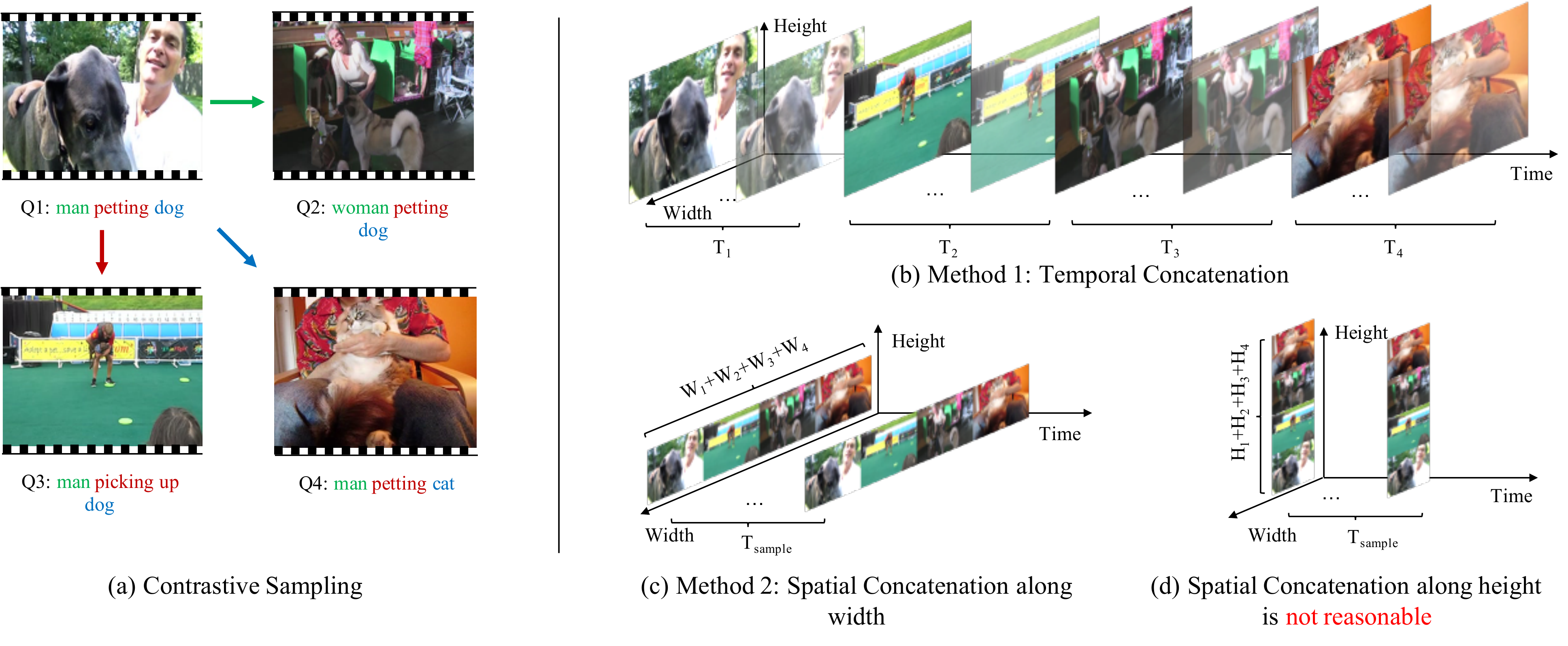}
    
    \caption{(a) illustrates contrastive sampling based on semantic roles. 
    Q1 contains a single agent (``man'') and a single patient (``dog''). 
    We use the SRL structure \ag{\dt{Arg0}}-\vb{\dt{Verb}}-\pat{\dt{Arg1}} but replace one queried object (Q2, Q4) or action (Q3).
    (b) shows temporal concatenation where we resize each video to the same width, height.
    (c) shows spatial concatenation where we resize the height and sample a fixed number of frames across the videos
    (d) shows an unreasonable spatial concatenation as videos have a top-down order (``ocean'' is always below ``sky'')
    }
    \label{fig:ds4_evl_stg}
\end{figure*}

\vspace{\sectionReduceTop}
\section{Method}
\vspace{\sectionReduceBot}
\label{sec:method}
We describe our sampling and concatenation process  which enables learning object relations for \tk (Section \ref{ss:bias_strats}), followed by details of \arch (Section \ref{ss:framework}) 
and relative position encoding scheme (Section \ref{ss:relpe}).
\subsection{Contrastive Sampling}
\label{ss:bias_strats}
\begin{table}[t]
\centering
\begin{tabular}{|c|c|c|c|c|}
\hline
Agent & Verb & Patient & Modifier & Instrument \\ \hline
Person & washes & cups & in a sink & with water. \\ \hline
\ag{\dt{Arg0}} & \vb{\dt{Verb}} & \pat{\dt{Arg1}} & \loc{\dt{ArgM-Loc}} & \inst{\dt{Arg2}} \\ \hline
\end{tabular}
\vspace{1.5mm}
\caption{An illustration of semantic-role assignment to a description. 
Here, the actor/agent (person) performs an action/verb (wash) using some instrument (water) at some location (sink).
}
\label{tab:srl_args}
\end{table}

Most large scale video datasets \cite{kay2017kinetics,caba2015activitynet,AbuElHaija2016YouTube8MAL} are curated from Internet sources like YouTube which rarely contain multiple instances of the same object in the same video.
\tk on such datasets can be trivially solved using object detection.

To mitigate this issue, we propose a two-step contrastive sampling method.
First, we assign semantic roles labels (SRLs) to every language descriptions of the videos
(see Table \ref{tab:srl_args})
and sample other descriptions by replacing each role in a one-hot style (Figure \ref{fig:ds4_evl_stg}(a)).

In the second step, we aggregate our samples.
One simple method is to present each video separately, similar to standard multiple-choice in Question-Answering tasks \cite{Zellers2018SWAGAL,Zellers2018FromRT,Lei2018TVQALC}; we call this ``separate'' (\sep) strategy (\ie the videos are viewed separately).
However, \sep doesn't force learning object relations, as one could independently add the scores for each referred object.
For instance, in Figure \ref{fig:ds4_evl_stg}-(a) we can score ``man'', ``petting'', ``dog'' individually and choose the objects in the video with the highest aggregate score essentially discarding object relations.

Alternatively, we generate new samples by concatenation along the time axis (\temp) or the width axis (\spat). 
For \temp, we resize the sampled videos to have the same width and height (Figure \ref{fig:ds4_evl_stg}(b)). 
For \spat, we resize the height dimension and uniformly sample $F$ frames for each video (Figure \ref{fig:ds4_evl_stg}(c)). 
Generally, it is not reasonable to concatenate along the height dimension as most real-world images obey up-down order (``sky'' is on the top while ``ocean'' is below) but not left-to-right order (Figure \ref{fig:ds4_evl_stg}(d)). 
Such concatenated videos, by construction, have multiple instances of the same object category.
To associate an instance described in the language query to its bounding box in the video, a model would need to disambiguate among similar object instances by exploiting their relations to the other objects.
For \eg, in Figure \ref{fig:ds4_evl_stg}(c) ``man'' or ``dog'' cannot be uniquely identified without considering other objects.
\begin{figure*}
    \centering
    \includegraphics[width=\linewidth]{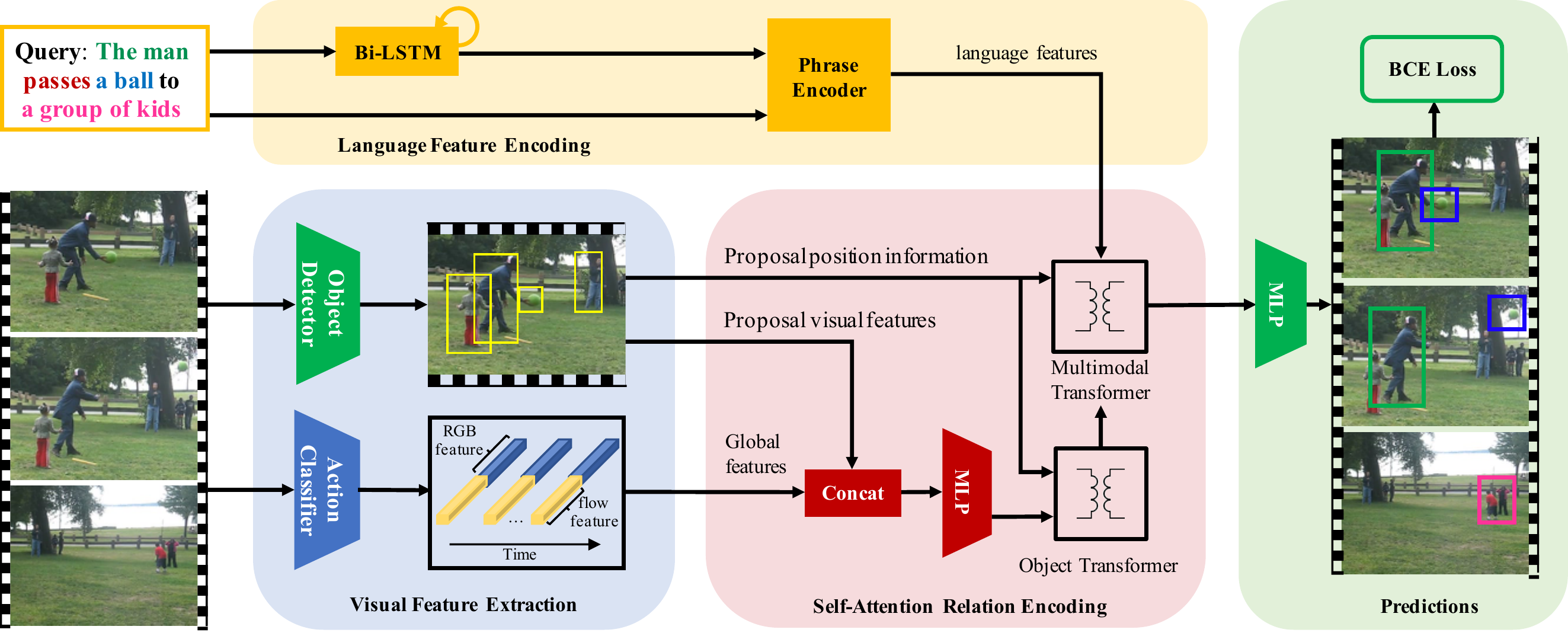}
    \caption{An overview of \arch. It takes a video-query pair as an input.
    A visual encoder extracts object features for each frame and concatenates them with segment features (rgb+flow).
    A language encoder encodes the whole query with a BiLSTM \cite{Schuster1997BidirectionalRN,hochreiter1997long} and then maintains a separate encoding for each phrase in the query (Eq.~\ref{eq: srl embed}).
    A Transformer \cite{vaswani2017attention} is first applied to the visual features to model object relations. 
    These self-attended visual features are fused  with the language features.
    Finally, a separate transformer models the interaction among the fused multi-modal features followed by a 2-layer MLP.
    \arch is trained with Binary Cross Entropy (BCE) loss.
    }
    \label{fig:framework}
\end{figure*}

\textbf{Caveats:}
(i) in \temp, one could use an activity proposal network like \cite{BSN2018arXiv,Gao2018CTAPCT} and bypass the problem altogether,
(ii) in \spat uniformly sampling $F$ frames from two different videos, would result in different parts of the image moving faster or slower and could partially affect our results.

\subsection{Framework}
\label{ss:framework}

\textbf{Notation:} We are given a video $V$ sampled with $F$ frames and a language description $L$ with $k$ roles. 
In general, not all $k$ roles can be visually grounded in the video, however, this information is not known apriori.
Given $P$ proposals for each frame using an object detector, we denote $O{=}\{p_{i,j}\}$ ($i^{th}$ proposal in $j^{th}$ frame) as the set of proposals in the video.
In \tk we learn the mapping $H:(V, O, L) \rightarrow [\{p^*_l\}_{j=1}^F]_{l=1}^k$ where $p^* \in O$.
That is, for each of the $k$ roles, we output a proposal $p^*$ in each frame.
We allow $p^*{=}\phi$ if the object is not visible in a particular frame, or the object cannot be localized.

We build a \arch framework that contains a Language Module to encode the query descriptions at the phrase level, a Visual Module to encode the object and frame level features in the video and a Multi-Modal Relation Module to encode both a language independent and dependent object relations.
Figure \ref{fig:framework} gives an overview of \arch.

\textbf{Language Module} first encodes the query $q=\{w_i\}_{i=1}^n$ as $n$ hidden vectors $[h_1, \hdots, h_n]$ with a Bi-LSTM \cite{hochreiter1997long,Schuster1997BidirectionalRN}.
The $j$-th Semantic Role Label (SRL) in query $q$, \texttt{Argj}, spanning a set of words $S_j$ (\emph{e.g.}, in Figure~\ref{fig:framework}, \texttt{Arg0} includes the words $S_0$ =  \{``The'', ``man''\}) is encoded as $\Tilde{q}_j$ is

\begin{equation}\label{eq: srl embed}
\Tilde{q}_j = \mathcal{M}_q(\mathcal{G}(\{\delta(w_i\in S_j)\cdot h_i\}_{i=1}^n))
\end{equation}
where $\delta(.)$ is an indicator function, and $\mathcal{G}(.)$ is an aggregation function. 
In VOGNet, we set $\mathcal{G}$ as the concatenation of the first word and the last word for each SRL, followed by $\mathcal{M}_q$ which denotes a Multiple Layer Perceptron (MLP).

\textbf{Visual Feature Extraction:}
An off-the-shelf object detector \cite{ren2015faster} returns $P$ proposals for each frame.
Let $p_{i,j}$ be the $i^{th}$ proposal in $j^{th}$ frame and $v_{i,j} {\in} \R^{d_v}$ be its ROI-pooled feature. 
Similarly, an action classifier returns temporal features containing image-level and flow-level features of the video. 
In general, the number of frames considered by the action classifier could be greater than $F$. 
We consider the local segment feature corresponding to the $F$ frames to get $s_j {\in} \R^{d_s}$, and append it to each proposal feature in $j^{th}$ frame. 
The final visual feature is $\hat{v}_{i,j} = \mathcal{M}_v(v_{i,j} || s_j)$, where $\mathcal{M}_v$ is a MLP.

\textbf{Object Transformer} is a transformer \cite{vaswani2017attention} and applies self-attention over the proposal features $\hat{v}_{i,j}$, \ie self-attention is applied to $P {\times} F$ proposals.
 We denote the self-attended visual features as $\hat{v}_{i,j}^{sa}$.
 Similar module is used in \cite{zhou2019grounded} but there are two differences: first, $\hat{v}_{i,j}$  contains additional segment features; second absolute positions 
 are replaced with relative position encoding (Section \ref{ss:relpe}).

\textbf{Multi-Modal Transformer:}
We concatenate the self-attended visual features $\hat{v}^{sa}$ and the language features $\Tilde{q}$ to get multi-modal features $m$ where 
$m[l,i,j] = [\hat{v}^{sa}_{i,j} || \Tilde{q}_l]$.
We apply self-attention with relative position encoding to get self-attended multi-modal features $m^{sa}$.
However, due to hardware limitations, it is extremely time consuming to perform self-attention over all the proposals especially when $P {\times} F {\times} k$ is large.
Thus, we perform this self-attention per frame \ie self-attention is applied to $P \times k$ features $F$ times.
Subsequently, $m^{sa}$ is passed through 2-layered MLP to get prediction for each proposal-role pair to get $\Tilde{m}^{sa}$.

\textbf{Loss Function:}
Let $L_g$ be the set of groundable roles \ie have a corresponding bounding box.
Thus, a proposal-role pair is considered correct if it has $IoU {\ge} 0.5$ and negative otherwise.
We train using Binary Cross Entropy (BCE) loss and average over the phrases with a bounding box:
\begin{align}
    L_{pred} = \frac{1}{|L_g|}\sum_{l_g \in L_g} \text{BCE}(\Tilde{m}^{sa}[l_g,i,j], gt[l_g,i,j])
\end{align}

\textbf{Minor changes for \sep}:
When training and evaluating models using \sep strategy we have access to the individual videos. 
Here, we use the temporal features to learn a \vb{\dt{Verb}} score which can be used to disambiguate between videos with the same objects but different verbs. 
In general, this didn't translate to other strategies and hence it is not included in our framework.

\subsection{Relative Position Encoding}
\label{ss:relpe}
Relative Position Encoding (RPE) uses the relative distances between two proposals as an additional cue for attention.
We denote the normalized positions of the proposal $p_{a,b}$ whose $5d$ coordinate is $[x_{tl}, y_{tl}, x_{br}, y_{br}, j]$ with $pos_{a,b} = [x_{tl}/W, y_{tl} / H, x_{br}/W, y_{br} / H, j/F]$.
We encode the relative distance between two proposals $A$ and $B$ as $\Delta_{A,B} = \mathcal{M}_p(pos_{A} - pos_{B})$, where $\mathcal{M}_p$ is a MLP.

Let the Transformer contain $n_l$ layers and $n_h$ heads.
Here, $\Delta_{A,B} \in \R^{n_h}$
When self-attention is applied to a batch 
 \begin{align}
    A(Q,K,V) = \text{SoftMax}({QK^T}/{\sqrt{d_k}})V
 \end{align}

We change this to
\begin{align}
    A(Q,K,V) = \text{SoftMax}((QK^T + \Delta[h])/{\sqrt{d_k}})V
\end{align} 

Note that $\Delta[h]$ has the same dimensions as $QK^T$ and leading to a simple matrix addition.
That is, our relative position encoding (RPE) encodes the distance between each proposal pair and this encoding is different for each head.
Intuitively, RPE biases the self-attention to weigh the contribution of other objects relative to their proximity. 

Our solution is based on prior work \cite{Shaw2018SelfAttentionWR} but differs in two key aspects: (i) the relative positions are not embedding layers rather modeled by a MLP to encode the difference
(ii) our relative encoding is different for different heads.
Another way to extend \cite{Shaw2018SelfAttentionWR} to visual setting would be to categorize the distances into multiple bins and learn encoding for each bin. 
We leave this study for future work.

\textbf{Caveat:}
While we resolve the issue of adding RPE to the transformer network efficiently, computation of $\Delta_{i,j}$ remains expensive as it requires $O(n^2)$ difference computation and is a bottleneck of our proposed solution.

\vspace{\sectionReduceTop}
\section{Experiments}
\vspace{\sectionReduceBot}
\label{sec:exp}
We briefly describe the dataset construction (see Appendix \ref{sec:sup_data_construct} for more details) followed by experimental setup, results and visualizations.

\subsection{Constructing ActivityNet-SRL}
\label{ss:data_const}
Our proposed dataset ActivityNet-SRL (ASRL) is derived from ActivityNet \cite{caba2015activitynet}, ActivityNet-Captions (AC) \cite{krishna2017dense} and ActivityNet-Entities (AE) \cite{zhou2019grounded}. 
There are two key steps in creating ASRL: 
(i) add semantic role labels (SRLs) to the descriptions in AC and filter it using heuristics
(ii) add lemmatized words for each groundable phrase labeled as a semantic role for efficient contrastive sampling.

For (i) we apply \cite{shi2019simple}, a BERT-based \cite{Devlin2019BERTPO} semantic-role labeling system to the video descriptions in AC.
We use the implementation provided in \cite{Gardner2017AllenNLP} trained on OntoNotes5\cite{Pradhan2013TowardsRL} which uses the PropBank annotation format \cite{palmer2005proposition}.
The obtained semantic-roles are cleaned using heuristics like removing verbs without any roles usually for ``is'', ``are'' etc.
In general, each description contains multiple ``verbs'' and we treat them separately.

\begin{table}[t]
    \centering
    \begin{tabular}{cccc}
\toprule
    \ag{\dt{Arg0}} & \pat{\dt{Arg1}} & \inst{\dt{Arg2}} & \loc{\dt{ArgM-Loc}} \\
\midrule
    42472 & 32455 & 9520 & 5082 \\
\bottomrule
    \end{tabular}\vspace{1.5mm}
    \caption{Number of annotated boxes in ASRL training set.}
    \label{tab:arg_stats}
\end{table}

For (ii) we utilize bounding box annotations in AE. 
First, we align the tokens obtained from the SRL system with the tokens of AE using \cite{spacy2}.
Then, for each phrase labeled with a semantic role, we check if the corresponding phrase in AE has a bounding box and mark the phrase as being groundable or not.
Since AE provides object names derived from the noun-phrases parsed using \cite{manning-etal-2014-stanford} we use them as the lemmatized word for the phrase.
Table \ref{tab:arg_stats} shows the top-4 semantic roles with bounding box annotations in the training set of ActivityNet-Entities.
We confine to this set of SRLs for contrastive sampling.

For training, we use the training set of ActivityNet which is the same as AC and AE.
However, to create test set for AE, we need the ground-truth annotations which are kept private for evaluative purposes.
As an alternative, we split the validation set of AE equally to create our validation and test set.
When contrastive sampling is used in training, we only sample from the train set.
However, since the size of validation and test sets is reduced, it is difficult to find 
contrastive examples. 
As a remedy, we allow sampling of contrastive examples from the test set during validation and vice versa for testing but never used in training.

\subsection{Dynamic Contrastive Sampling}
\label{ss:ecs}
While Contrastive Sampling is mainly used to create the validation and test sets to evaluate \tk, it can also be used for training where speed is the bottleneck.
Given a particular description belonging to training index $T$ containing roles $R = [r_1, \hdots, r_k]$ with the corresponding lemmatized words $S = [s_1, \hdots, s_k]$ we need to efficiently sample other descriptions with the same semantic-roles but containing one different lemmatized word. 
That is, we need to sample indices $T_i$ whose lemmatized words are $S_i = [s_1, \hdots, s_i', \hdots s_k]$ for every $1\le i \le k$.

To address this, we first create a separate dictionary $D_{i}$ for each semantic role $r_i$ containing a map from the lemmatized words to all the annotation indices where it appears as $r_i$.
Given $S$, we can efficiently obtain $T_i$ by randomly sampling from the set $E_i = \cap_{j \in \{1...k\}, j \neq i}D_j(s_j)$.

Due to hardware limitations, we restrict $k \le 4$.
For $k > 4$, we randomly drop $k-4$ indices.
If $k < 4$, then we randomly sample a training index $T_j$ with the only restriction that the $T$ and $T_j$ describe different videos.

\subsection{Experimental setup}
\label{ss:evlmtr}

\textbf{Dataset Statistics:} 
In total, ASRL contains $39.5k$ videos with $80k$ queries split into training, validation, and testing with $31.7k, 3.9k, 3.9k$ videos and $63.8k, 7.9k, 7.8k$ queries.
Each video contains around $2$ queries containing $3.45$ semantic roles and each query has around $8$ words. 

\textbf{Evaluation Metrics:}
We compute the following four metrics:
(i) \textbf{accuracy}: correct prediction for a given object in a query (recall that a query has references to multiple objects)
(ii) \textbf{strict accuracy}: correct prediction for all objects in the query
(iii) \textbf{consistency}: the predictions for each object lie in the same video
(iv) \textbf{video accuracy}: predictions are consistent and lie in the correct video.
While strict accuracy is the most important metric to note for \tk, other metrics reveal useful trends helpful for model diagnosis and building robust \tk models and datasets. 

\textbf{Metric Computation:} 
In AE, the noun phrases are only localized in the frame where it is most easily visible.
This complicates the evaluation process when the same objects appear across multiple frames (a common occurrence).
Thus,  we select the highest-scoring proposal box for each role in the query in every frame and set a score threshold.
Given a phrase referring to a groundable object, we consider the prediction correct when the predicted box in an annotated frame has an $IoU \ge 0.5$ with a ground-truth box.
This allows us to compute accuracy in a single video single query (\svsq) setting.

For \sep, \temp, \spat we have additional information about which video frames and proposal boxes are not ground-truths.
To evaluate \sep: we check if the predicted video is correct (which gives us video accuracy), and if so compute the accuracy similar to \svsq.

In \temp and \spat, for a given role if the predicted boxes not belonging to the ground-truth video have a score higher than a threshold, then the prediction for the role is marked incorrect. 
If the boxes are in the ground-truth video, we evaluate it similar to \svsq (see Appendix \ref{sec:sup_eval} for examples of each strategy).

\begin{table*}[t]
\resizebox{\textwidth}{!}{%
\centering
\begin{tabular}{c|c|cc|ccc|cccc|cccc}
\tpr
 \multirow{2}{*}{} & \multirow{2}{*}{Model} & \multicolumn{2}{c|}{\svsq} & \multicolumn{3}{c|}{\sep} & \multicolumn{4}{c|}{\temp}      & \multicolumn{4}{c}{\spat}      \\
 &      & \acc         & \sacc       & \acc    & \vacc   & \sacc  & \acc   & \vacc  & \cons  & \sacc  & \acc   & \vacc  & \cons  & \sacc  \\
 \midrule
\multirow{3}{*}{\gt} & \bsi & 75.31 & 56.53 & 39.78 & 51.14 & 30.34 & 17.02 & 7.24 & 34.73 & 7.145 & 16.93 & 9.38 & 49.21 & 7.02 \\
 & \bsv  & 75.42       & 57.16      & 41.59  & 54.16  & 31.22 & 19.92 & 8.83  & 31.70 & 8.67  & 20.18 & 11.39 & 49.01 & 8.64  \\
 & \arch & \textbf{76.34}       & \textbf{58.85}      & \textbf{42.82}  & \textbf{55.64}  & \textbf{32.46} & \textbf{23.38} & \textbf{12.17} & \textbf{39.14} & \textbf{12.01} & \textbf{23.11} & \textbf{14.79} & \textbf{57.26} & \textbf{11.90} \\

\midrule
\multirow{3}{*}{\phun} & 
 \bsi  & \textbf{55.22} & \textbf{32.7} & 26.29 & 46.9  & 15.4  & 9.71  & 3.59 & 22.97 & 3.49 & 7.39 & 4.02 & \textbf{37.15} & 2.72 \\
 & 
 \bsv  & 53.30  & 30.90 & 25.99 & 47.07 & 14.79 & 10.56 & 4.04 & \textbf{29.47} & 3.98 & 8.54 & 4.33 & 36.26 & 3.09 \\
 &  
 \arch & 53.77 & 31.9 & \textbf{29.32} & \textbf{51.2}  & \textbf{17.17} & \textbf{12.68} & \textbf{5.37} & 25.03 & \textbf{5.17} & \textbf{9.91} & \textbf{5.08} & 34.93 & \textbf{3.59} \\

\btr
\end{tabular}%
}
\vspace{1.5mm}
\caption{
Comparison of \arch against \bsi and \bsv.
\gt and \phun use $5$ and $100$ proposals per frame.
Here, \acc: Grounding Accuracy, \vacc: Video accuracy, \cons: Consistency, \sacc: Strict Accuracy (see Section \ref{ss:evlmtr} for details). On the challenging evaluation metrics of \temp and \spat,  \arch (ours) shows significant improvement over competitive image and video grounding baselines.
}
\label{tab:main_cmp}
\end{table*}

\begin{table}[t]
\centering
\resizebox{\linewidth}{!}{%
\begin{tabular}{c|cc|cc|cc}
\tpr
     & \multicolumn{2}{c|}{\svsq} & \multicolumn{2}{c|}{\temp} & \multicolumn{2}{c}{\spat} \\
     & \acc         & \sacc       & \acc         & \sacc       & \acc         & \sacc       \\
\midrule
\svsq & 76.38       & 59.58      & 1.7         & 0.42       & 2.27        & 0.6        \\
\temp & 75.4        & 57.38      & 23.07       & 12.06      & 18.03       & 8.16       \\
\spat & 75.15       & 57.02      & 22.6        & 11.04      & 23.53       & 11.58     \\
\btr
\end{tabular}%
}
\vspace{1.5mm}
\caption{Evaluation of \arch in \gt setting by training (first column) and testing (top row) on \svsq, \temp, \spat respectively}
\label{tab:st_cross}
\end{table}
\textbf{Baselines:} 
Prior work on \tk cannot be evaluated on ASRL due to their restrictive formulations.
For instance, \cite{Zhou2018WeaklySupervisedVO} grounds all objects when using \temp and \spat resulting in $0$ accuracy and \cite{Chen2019WeaklySupervisedSG} needs spatio-temporal tubes.

Recently, \cite{zhou2019grounded} proposed GVD, a model for grounded video description. 
GVD calculates its grounding accuracy by feeding the ground-truth description into a  captioning system and finding the highest scored objects.
However, this is not applicable to our task because it considers the language in a sequential manner.
For an input query ``man throwing ball'', GVD would ground ``man'' without looking at the remaining description and thus fail at grounding in our proposed contrastive setting.

As an alternative, we propose two competitive baselines:
(i) \bsi: an image grounding system which treats each frame independently and does not explicitly encode object relations.
(ii) \bsv: a video grounding system based on GVD using an object transformer to encode object relations.
For fair comparisons, we use the same language features, visual features (the proposal and segment features) for both \bsi and \bsv 

\textbf{Implementation details:} 
We re-use the extracted visual features  provided by \cite{zhou2019grounded} for AE.
The object proposals and features are obtained from a FasterRCNN \cite{ren2015faster} trained on visual genome \cite{krishna2017visual}.
Segment features (both RGB and Flow features) are obtained using TSN \cite{wang2016temporal} trained on ActivityNet \cite{caba2015activitynet}.
For each video, $F{=}10$ frames are uniformly sampled and for each frame, we consider $P{=}100$ proposals which gives a recall of $88.14\%$.
However, training with $100$ proposals is time-consuming and computationally expensive.
Instead, we introduce \gt setting where we use exactly $5$ proposals per frame. 
In unannotated frames, it includes the highest-scoring proposals, and for annotated frames, for each ground-truth box, it prioritizes the proposal having the highest $IoU$.
\gt maintains a similar recall score ($86.73\%$), and allows experimenting with more variations and sets upper performance bound.

For self-attention, both Object Transformer (OTx) and Multi-Modal Transformer (MTx) use multi-head attention \cite{vaswani2017attention} with $n_l{=}1$ layer and $n_h{=}3$ heads unless mentioned otherwise.
In general, Object Transformer (OTx) applies self-attention across all proposals and frames whereas the Multi-Modal Transformer (MTx) applies self-attention to each frame separately due to higher computation load.
We train all models until the validation accuracy saturates.
For \sep, \temp, \spat we found 10 epochs with batch size $4$ for \gt and $2$ for \phun, using Adam with learning rate $1e^{-4}$ to be sufficient for most models.
For \svsq, we set batch size $4$ for all models.
We use the model with the highest validation accuracy for testing.
We set the threshold used in evaluating \temp and \spat as $0.2$ for \gt and  $0.1$ for \phun across all models.
More implementation details are provided in Appendix \ref{sec:sup_bsl}.

\begin{table}[t]
\centering
\begin{tabular}{cc|cc|c|c}
\tpr
 &  & \multicolumn{2}{c|}{\sep} & \temp & \spat \\
Train & Test & \acc & \vacc & \acc & \acc \\
\midrule
Rnd    & CS  & 44.9  & 57.6  & 22.89 & 22.72 \\
CS+Rnd & CS  & 44.8  & 56.94 & 23.07 & 23.53 \\
CS+Rnd & Rnd & 57.44 & 74.1  & 36.48 & 36.05 \\
\btr
\end{tabular}\vspace{1.5mm}
\caption{
Comparison of Contrastive Sampling (CS) vs Random Sampling (Rnd) for training (row-1,2) and evaluation (row-2,3). 
}
\label{tab:cs_rand}
\end{table}

\begin{table}[t]
\centering
\begin{tabular}{cc|cccc}
\tpr
\#vids & \#epochs & \acc & \vacc & \cons & \sacc \\
\midrule
2 & 20 & 20.18 & 10.18 & 52.45 & 8.84  \\
3 & 13 & 21.7  & 13.33 & 55.55 & 10.68 \\
5 & 8  & 23.34 & 14.53 & 56.51 & 11.71 \\
\btr
\end{tabular}\vspace{1.5mm}
\caption{Training \arch in \spat setting with different number of concatenated videos and tested on \spat with $4$ videos.}
\label{tab:nvids}
\end{table}

\begin{table}[t]
\centering
\resizebox{\linewidth}{!}{%
\begin{tabular}{l|cccc}
\tpr
\multicolumn{1}{c}{\spat} & \acc & \vacc & \cons & \sacc \\
\midrule
\bsi             & 17.03 & 9.71  & 50.41 & 7.14  \\
\quad +OTx(1L, 3H)   & 19.8  & 10.91 & 48.34 & 8.45  \\
\qquad +RPE            & 20.2  & 11.66 & 49.21 & 9.28  \\
\quad +MTx(1L, 3H)   & 19.23 & 10.49 & 48.19 & 8.14  \\
\qquad +RPE            & 19.09 & 10.46 & 50.09 & 8.23  \\
\quad +OTx(3L, 6H)   & 21.14 & 12.1  & 49.66 & 9.52 \\ \quad +OTx + MTx + RPE & \textbf{23.53} & \textbf{14.22} & \textbf{56.5}  & \textbf{11.58} \\
\midrule
\arch & & & & \\
\quad +MTx(3L,6H)    & 24.24 & \textbf{15.36} & 57.37 & 12.52 \\
\qquad +OTx(3L,6H)     & \textbf{24.99} & 7.33  & \textbf{66.29} & \textbf{14.47} \\
\btr
\end{tabular}%
}
\vspace{1.5mm}
\caption{Ablative study comparing gains from Multi-Modal Transformer (MTx) and Object Transformer (OTx) and Relative Position Encoding (RPE). L: Number of Layers, H: Number of Heads in the Transformer. Note that \arch = \bsi +MTx(1L,3H) +OTx(1L,3H) + RPE }
\label{tab:mabl}
\end{table}

\subsection{Results and Discussions}
In Table \ref{tab:main_cmp}, we compare \arch against two baselines \bsi and \bsv across \gt ($5$ proposal boxes per frame) and \phun ($100$ proposal boxes per frame).

\textbf{Comparison of Strategies:}
We note that in the \svsq column, all the models perform comparably.
However, these results fail to generalize to other cases which indicates that evaluating on \svsq is insufficient.
Next, the \sep column
shows that models can distinguish contrastive samples by considering the contribution of each object independently with very high accuracy and can easily distinguish similar examples achieving $\approx50\%$ on video accuracy even in the \phun setting.
Such cues are not present in \spat and \temp where the model is given a single video and single query but now the video contains more than one actor performing some action.
The performance in both \spat and \temp is still very low (strict accuracy for \phun is ${<}5\%$),
which suggests that \tk remains an extremely challenging problem for current state-of-art models.
\begin{figure*}
    \centering

    \includegraphics[width=\linewidth]{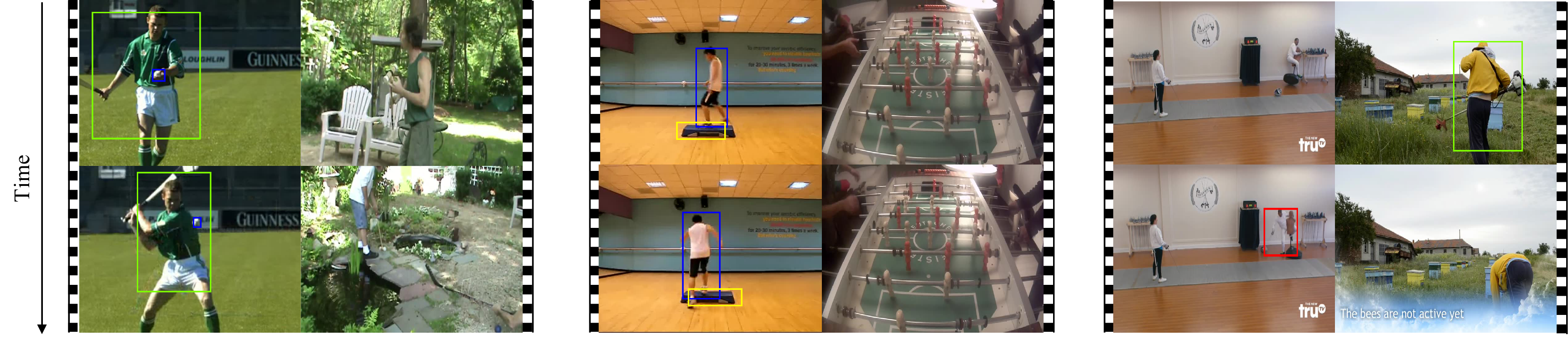}
        \caption{Left(L): concatenated using \spat with query: [\ag{\dt{Arg0}: The man}] [\vb{\dt{Verb}: throws}] [\pat{\dt{Arg1}: the ball}] [\inst{\dt{Arg2}: in the air}].
        Middle(M): concatenated using \temp with query: [\pat{\dt{Arg1}: He}] [\vb{\dt{Verb}: spins}] [\adir{\dt{ArgM-DIR}: around the board}].
        Right(R): concatenated using \spat with query:  [\ag{\dt{Arg0}: He}] [\vb{\dt{Verb}: attaches}] [\pat{\dt{Arg1}: a dummy}].
        In L, R the top-two and bottom-two frames are seen concatenated. 
        In M, there are four frames following the order: tl-bl-tr-br.
        In L,M our model \arch correctly finds the referred objects (``man'', ``ball'', ``boy'', ``board''). 
        In R: \arch is unable to find ``dummy'' and ends up localizing the incorrect person.
        }
        \vspace{-2mm}
    \label{fig:vis}
\end{figure*}

\textbf{Comparison with Baselines:} 
For both \temp and \spat, we find \bsi performs relatively well (${\approx}17\%$ in \gt) despite not using any object relations.
This is likely because the model can exploit attribute information (such as ``red shirt'') in the phrases.
\bsv which uses language independent object relations obtains gains of $2{-}3\%$.
Finally, \arch, which additionally uses language-dependent object relations, outperforms \bsv by another $3{-}4\%$.

\textbf{\gt vs \phun:}
We observe that both \gt and \phun
follow similar patterns across metrics suggesting \gt is a good proxy to explore more settings. 
For the remaining experiments, we consider only the \gt setting.

\textbf{Performance across Strategies}:
Table \ref{tab:st_cross} shows that  \arch trained in \spat and \temp settings performs competitively on \svsq (maintaining ${\approx}75\%$ accuracy).
However, the reverse is not true \ie models trained on \svsq fail miserably in \spat and \temp (accuracy is ${<}3\%$).
This suggests that both \temp and \spat moderately counter the bias caused by having a single object instance in a video.
Interestingly, while \arch trained on \temp doesn't perform well on \spat (performance is worse than \bsv trained on \spat), when \arch is trained on \spat and tested on \temp it significantly outperforms \bsv trained in \temp.
This asymmetry is possibly because the multi-modal transformer is applied to individual frames.

\textbf{Contrastive Sampling}: 
Table \ref{tab:cs_rand} compares Contrastive Sampling (CS) to a Random Sampling (RS) baseline for evaluation and training.
Using RS for validation, \sep video accuracy is very high $75\%$ implying that CS is a harder case; similarly, we find higher performance in both \temp and \spat cases.
Interestingly, using RS for training is only slightly worse for \spat, \temp while outperforming in \sep.
Thus, CS in \spat and \temp helps learn better object relations, but random sampling remains a very competitive baseline for training.
Table \ref{tab:nvids} 
shows that using more videos in training helps; we use $4$ videos due to GPU memory considerations and training time.

\textbf{Ablation Study:}
In Table \ref{tab:mabl} we record the individual contributions of each module in \spat.
We observe: 
(i) self-attention via object is an effective way to encode object relations across frames 
(ii) multi-modal transformer applied on individual frames gives modest gains but falls short of object transformer due to lack of temporal information 
(iii) relative position encoding (RPE) boosts strict accuracy for both transformers 
(iv) object transformer with 3 layers and 6 heads performs worse than using a single multi-modal transformer \ie adding more layers and attention heads to object transformer is not enough
(v) using both object and multi-modal transformers with more layers and more heads gives the best performing model.

\subsection{Visualizations}
For qualitative analysis, we show the visualizations of \spat and \temp strategies in Figure \ref{fig:vis}.
In the interest of space, we use $k{=}2$ contrastive sampling (visualizations with $k{=}4$ are provided in the Appendix \ref{sec:sup_vis}).
In the first image, the videos are concatenated along the width axis and both contain a ``man'' and ``ball''. 
Our model correctly identifies which ``ball'' is being thrown into the air and by whom.
Note that only viewing the last frame doesn't uniquely identify if the ``man'' visible in the current frame has thrown the ball.
In general, our \spat model performed with high consistency \ie it chose objects nearer to each other which we attribute to RPE.
In the second image, the videos are concatenated along the time-axis and in both videos, the person ``spins'' something. 
Using ``board'' as an additional cue, our model correctly finds both ``the person'' and the ``board that he spins''.
Our \temp model performs slightly worse than \spat model possibly because encoding temporal information is more challenging.
Finally, in the third image, our model grounds ``he'' incorrectly likely due to not being able to ground ``dummy''.

\vspace{\sectionReduceTop}
\section{Conclusion}
\vspace{\sectionReduceBot}
\label{sec:conclusion}

In this work, we analyze the problem of \tk which aims to localize the referred objects in a video given a language query.  
We show that semantic-role labeling systems can be used to sample contrastive examples. 
We then enforce that the model views the contrastive samples as a whole video so that the model explicitly learns object relations.
We further propose an additional self-attention layer to capture language dependent object relations along with a relative position encoding.
Finally, we validate our proposed model \arch on our dataset \acs which emphasizes the role of object interactions.\\ \\

\small{
\textbf{Acknowledgement:} We thank the anonymous reviewers for their suggestions.
This research was supported, in part, by the Office of Naval Research under grant \#N00014-18-1-2050.
}

\appendix
\begin{center}
  {\large \bf Appendix \par}
\end{center}

\numberwithin{equation}{section}
\setcounter{table}{0}
\setcounter{figure}{0}
\setcounter{equation}{0}
Appendix provides details on: 
\begin{enumerate}
    \itemsep0em
    \item Relative Position Encoding (Section \ref{sec:sup_rpe_det})
    \item \acs construction and statistics (Section \ref{sec:sup_data_construct})
    \item Evaluation Metrics (Section \ref{sec:sup_eval})
    \item Implementation Details (Section \ref{sec:sup_bsl})
    \item Additional Experiments (Section \ref{sec:sup_expts})
    \item Visualizations (Section \ref{sec:sup_vis})
\end{enumerate}

\vspace{\sectionReduceTop}
\section{Relative Position Encoding Discussion}
\vspace{\sectionReduceTop}
\label{sec:sup_rpe_det}
In this section, we describe the challenges of using relative position encoding, followed by an overview of the method used in \cite{Shaw2018SelfAttentionWR} and finally show how we adapt their formulation to our setting.
For an overview of the technical details of the Transformer \cite{vaswani2017attention}, we refer to
the following well-written blogs ``The Annotated Transformer''\footnote{https://nlp.seas.harvard.edu/2018/04/03/attention.html}, ``The Illustrated Transformer''\footnote{http://jalammar.github.io/illustrated-transformer/}, ``Transformers From Scratch''\footnote{http://www.peterbloem.nl/blog/transformers}.

In general, Transformer performs self-attention with multiple heads and multiple layers. 
For a particular head, to compute self-attention, it derives the query $Q$, key $K$ and value $V$ from the input $X$ itself as follows:
\begin{align}
    Q = W_qX \quad K = W_kX \quad V = W_vX 
\end{align}
Using the derived $Q,K,V$ triplet, it assigns new values to each input $X$ using attention $A$ given by
\begin{align}
    A(Q,K,V) &= \text{SoftMax}\left(\frac{QK^T}{\sqrt{d_z}}\right) V
\end{align}
Here $Q, K, V$ are each of shape $B \times T \times d_z$ where $B$ is the batch size, $T$ is the sequence length, and $d_z$ is the dimension of each vector.
The attention $A$ can be computed efficiently using batch matrix multiplication since the multiplication $QK^T$ and the subsequent multiplication with $V$ have the common $B \times T$.
For instance, when computing $QK^T$ we perform batch matrix multiplication with $B \times T \times d_z$ and $B \times d_z \times T$ resulting in $B$ matrix multiplications to give $B \times T \times T$ matrix.

Since the attention mechanism itself doesn't encode the positions of the individual $T$ vectors, it is insensitive to the order of the $T$ inputs. 
To address this, a position encoding is added to each of the $T$ inputs to make the transformer dependent on the order of inputs.
\cite{Shaw2018SelfAttentionWR} follows up by using an additional relative position encoding. 
They define two new matrices $a_{i,j}^K$and $a_{i,j}^V$ (both of shape $B \times T \times T \times d_z$) and change the attention equation as follows:
\begin{align}
    \label{eqn:not_rpe}
    A(Q,K,V) &= \text{SoftMax}\left(\frac{Q(K^T + a_{i,j}^K)}{\sqrt{d_z}}\right) (V + a_{i,j}^V)
\end{align}

As \cite{Shaw2018SelfAttentionWR} notes, this removes the computation efficiency in the original transformer due to computation of $a_{i,j}^K$ for all pairs, and more importantly, the efficient batch matrix multiplication cannot be used due to addition of $a_{i,j}^K$ to $K$ making it of shape $B \times T \times T \times d_z$.
To resolve this, they propose the following equivalent formulation for computing $QK^T$ (similarly for multiplying $V$):
\begin{align}
    Q(K^T + a_{i,j}^K) &= QK^T + Qa_{i,j}^K
\end{align}
Such formulation removes the additional time to compute $K+a_{i,j}^K$ which would otherwise be a major bottleneck.

There are two related challenges in adopting it to the visual domain: 
(i) the positions are continuous rather than discrete
(ii) both $a_{i,j}^K$ and $a_{i,j}^V$ have $d_z$ dimension vector which is highly over-parameterized version of the $5d$ position vector ($d_z \gg 5$).
To address (i) we use a $\mathcal{M}_p$ (MLP) to encode the $5d$ position which is a reasonable way to encode continuous parameters.
For (ii) we change Eq.~\ref{eqn:not_rpe} as
\begin{align}
    A(Q,K,V) &= \text{SoftMax}\left(\frac{QK^T + \Delta}{\sqrt{d_z}}\right)V 
\end{align}
Here $\Delta$ is of shape $B \times T \times T$ same as $QK^T$ and $\Delta$ is computed from the relative positions of two object proposals $p_i, p_j$ as $\Delta_{i,j} = \mathcal{M}_p(p_i - p_j)$ is a scalar.
For added flexibility, we have that $\Delta_{i,j} \in \R^{n_h}$ where $n_h$ is the number of heads allowing us to use different $\Delta$ for different heads.

As mentioned in Section 3.3 (of the main paper), the computation of $p_i - p_j$ for every pair remains the major bottleneck of our proposed relative position encoding.

\vspace{\sectionReduceTop}
\section{Dataset Construction}
\vspace{\sectionReduceTop}
\label{sec:sup_data_construct}
\begin{table*}[t]
\centering
\begin{tabular}{ccc}
\toprule

\multicolumn{3}{c}{Sentence: A woman is seen speaking to the camera while holding up various objects and begins brushing her hair.} \\
\midrule
\vb{\dt{Verb}} & Semantic-Role Labeling & Considered Inputs\\
\midrule
is & \begin{tabular}[c]{@{}c@{}}A woman \vb{{[}V: is{]}} seen speaking to the camera while \\ holding up various objects and begins brushing her hair.\end{tabular} & x \\
\midrule
seen & \begin{tabular}[c]{@{}c@{}}A woman is \vb{{[}V: seen{]}} \pat{{[}ARG1: speaking to the camera} \\ \pat{while holding up various objects and begins brushing her hair{]}} \end{tabular} & x \\
\midrule
speaking & \begin{tabular}[c]{@{}c@{}}\ag{{[}ARG0: A woman{]}} is seen \vb{{[}V: speaking{]}} \inst{{[}ARG2: to the camera{]}} \\ \atmp{{[}ARGM-TMP: while holding up various objects{]}} \\ and begins brushing her hair .\end{tabular} & 
\begin{tabular}[c]{@{}c@{}}\ag{A woman} \vb{speaking} \inst{to the camera} \\ \atmp{while holding up various objects}\end{tabular} \\
\midrule
holding & \begin{tabular}[c]{@{}c@{}}\ag{{[}ARG0: A woman{]}} is seen speaking to the camera while \\ \vb{{[}V: holding{]}} \adir{{[}ARGM-DIR: up{]}} \pat{{[}ARG1: various objects{]}} \\ and begins brushing her hair .\end{tabular} & 
\begin{tabular}[c]{@{}c@{}}\ag{A woman} \vb{holding} \adir{up} \\ \pat{various objects}\end{tabular} \\
\midrule
begins & \begin{tabular}[c]{@{}c@{}}\ag{{[}ARG0: A woman{]}} is seen speaking to the camera while holding \\ up various objects and \vb{{[}V: begins{]}} \pat{{[}ARG1: brushing her hair{]}} \end{tabular} & 
x \\
\midrule
brushing & \begin{tabular}[c]{@{}c@{}}\ag{{[}ARG0: A woman{]}} is seen speaking to the camera while holding \\ up various objects and begins \vb{{[}V: brushing{]}} \pat{{[}ARG1: her hair{]}} \end{tabular} & 
\begin{tabular}[c]{@{}c@{}}\ag{A woman} \vb{brushing} \\ \pat{her hair}\end{tabular}\\

\bottomrule
\end{tabular}
\vspace{1.5mm}
\caption{An example of applying semantic role labeling to the video description. Each verb is treated independent of each other and the verbs ``is'', ``seen'', ``begins'' are not considered. For all other verbs, the last column shows the considered input to the system}
\label{tab:sup_srl_eg}
\end{table*}

We derive \acs from ActivityNet-Entities (AE) \cite{zhou2019grounded} and ActivityNet-Captions (AC) \cite{krishna2017dense} (Section \ref{ss:sup_asrl_const}), provide the train, valid, and test split construction and statistics (Section \ref{ss:sup_tvt_split}), show the distribution of the dataset (Section \ref{ss:sup_ds_dist}) and finally compare ActivityNet against other datasets with object annotations (Section \ref{ss:sup_ds_choice}).

\subsection{Constructing ASRL}
\label{ss:sup_asrl_const}

We first use a state-of-the-art BERT \cite{Devlin2019BERTPO} based semantic role labeling system (SRL) \cite{shi2019simple} to predict the semantic roles of the video descriptions provided in AC. 
For SRL system, we use the implementation provided in AllenNLP \cite{Gardner2017AllenNLP} \footnote{see https://demo.allennlp.org/semantic-role-labeling for a demo}.
It is trained on OntoNotes 5 \cite{Pradhan2013TowardsRL} which uses PropBank annotations \cite{palmer2005proposition}.
PropBank annotations are better suited for \vb{\dt{Verb}} oriented descriptions. 
The system achieves $86.4\%$ on OntoNotes5. 
To ensure the quality, we randomly picked 100 samples and looked at the various labeled roles. 
We found a majority of these to be unambiguous and satisfactory. 
The few that were not found were removed by the following heuristics: 
(i) in a sentence like ``Man is seen throwing a ball'', we remove the ``seen'' verb even though it is detected because ``seen'' verb doesn't provide any extra information
(ii) similarly we also remove single verbs like ``is'', ``was'' which are already considered when some other verb is chosen
(iii) finally, in a small number of cases, no semantic-roles could be found, and such cases were discarded.
In general, each description can contain multiple verbs, in such cases, we treat each verb separately.
Table \ref{tab:sup_srl_eg} shows this with an example.

\begin{figure}
    \centering
    \includegraphics[width=\linewidth]{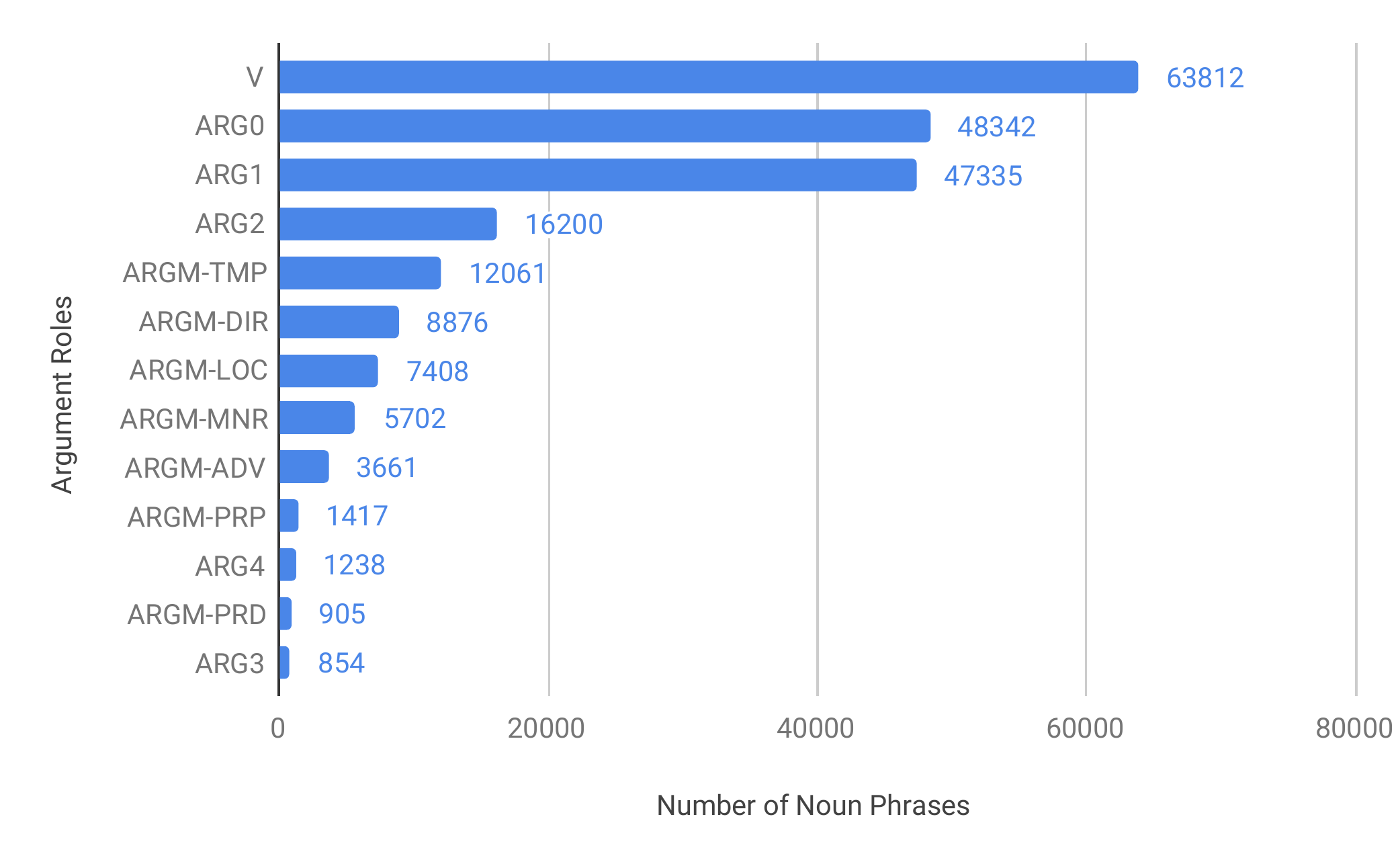}
    \caption{Number of Noun-Phrases for each role}
    \label{fig:sup_obj_np}
\end{figure}

\begin{figure}
    \centering
    \includegraphics[width=\linewidth]{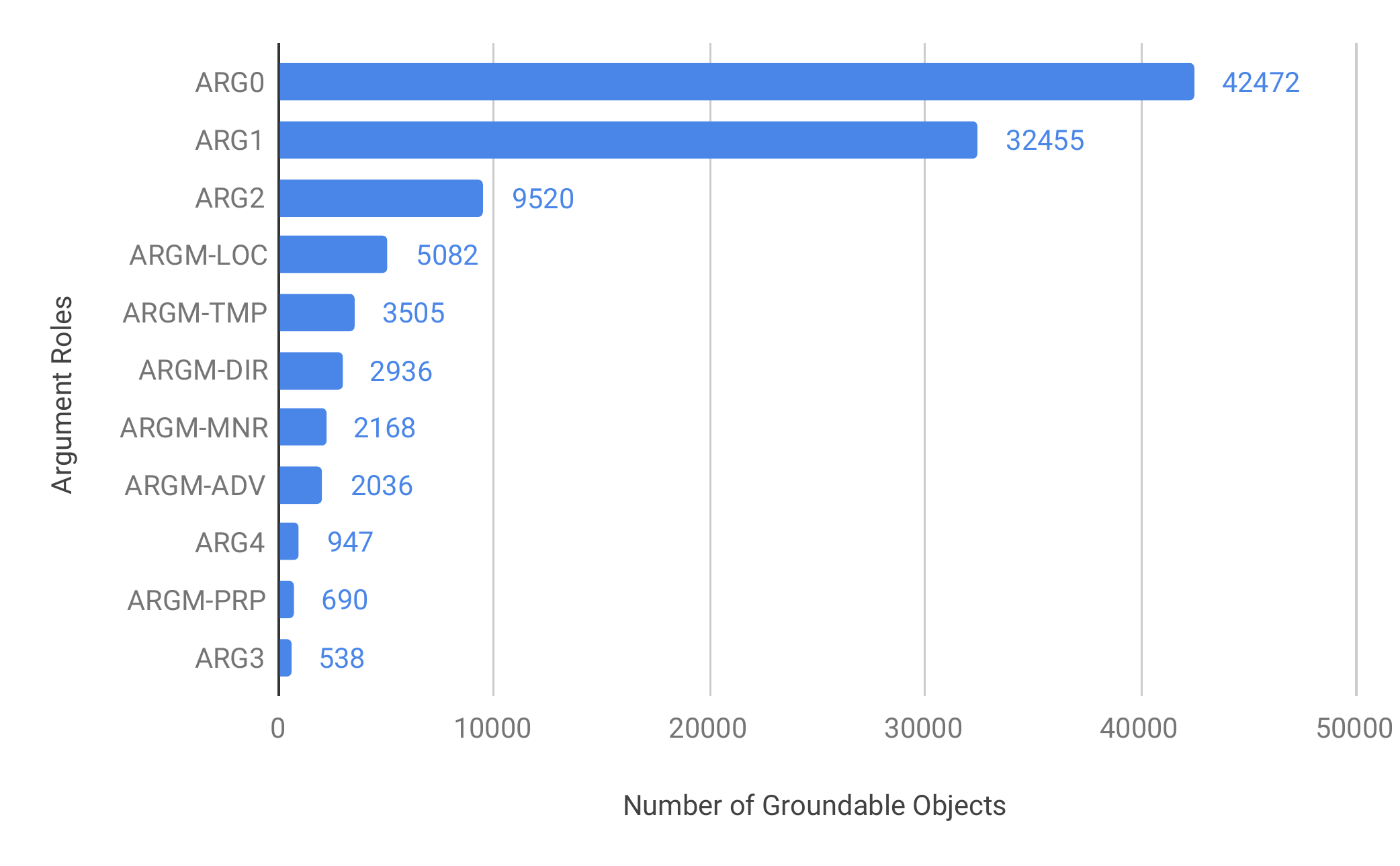}
    \caption{Number of groundable objects for each role
    }
    \label{fig:sup_grnd_obj_role}
\end{figure}

Once we have all the SRL annotated, we align them with the annotations of AE. 
This is non-trivial due to mis-match between the tokenization used by AE (which is based on Stanford Parser \cite{manning-etal-2014-stanford}) compared to the tokenization used in AllenNLP \cite{Gardner2017AllenNLP}.
Thus, we utilize the Alignment function provided in spacy v2 \cite{spacy2} to align the tokens from the two systems. 
To provide bounding box information to each role, we look at the tokens within the boundaries of the semantic role, and if any of them has been assigned a bounding box, we mark the semantic-role groundable, and assign it the corresponding bounding box.
Figure \ref{fig:sup_obj_np} shows the most common considered roles followed by Figure \ref{fig:sup_grnd_obj_role} depicting the most common roles which have a bounding box annotations (groundable roles).
Note that a particular role could be considered multiple times, \eg in Table \ref{tab:sup_srl_eg} ``A woman'' is considered for each of the verbs ``speaking'', ``holding'', ``begins'', ``brushing'' \ie some of the roles (in particular \ag{\dt{Arg0}}) are counted more than once.
While some roles like \atmp{\dt{ArgM-TMP}} and \adir{\dt{ArgM-DIR}} appear more often than \loc{\dt{ArgM-LOC}} (see Figure \ref{fig:sup_obj_np}), the number of groundable instances for the latter is much higher as locations are generally easier to localize. 
Further, note that \vb{\dt{Verb}} doesn't refer to an object and hence doesn't have any corresponding bounding boxes.

After having matched the annotated semantic roles with the bounding box annotations from AE, we lemmatize the arguments and create a dictionary for efficient contrastive sampling (as described in Section 4.2 in the main paper).
To obtain the lemmatized words, we use the object-name annotations given in AE which are themselves derived from stanford parser \cite{manning-etal-2014-stanford}.
To lemmatize the verbs, we use the inbuilt lemmatizer in spacy \cite{spacy2}.

\subsection{Training, Validation, Test Splits}
\label{ss:sup_tvt_split}

Once the roles and lemmatized words have been assigned, we need to create a train, validation and test splits. 

\textbf{Train:} We keep the same train split as AC, AE, ActivityNet.
This allows using activity classification networks like TSN \cite{wang2016temporal} trained on ActivityNet.

\begin{figure}
    \centering
    \includegraphics[width=\linewidth]{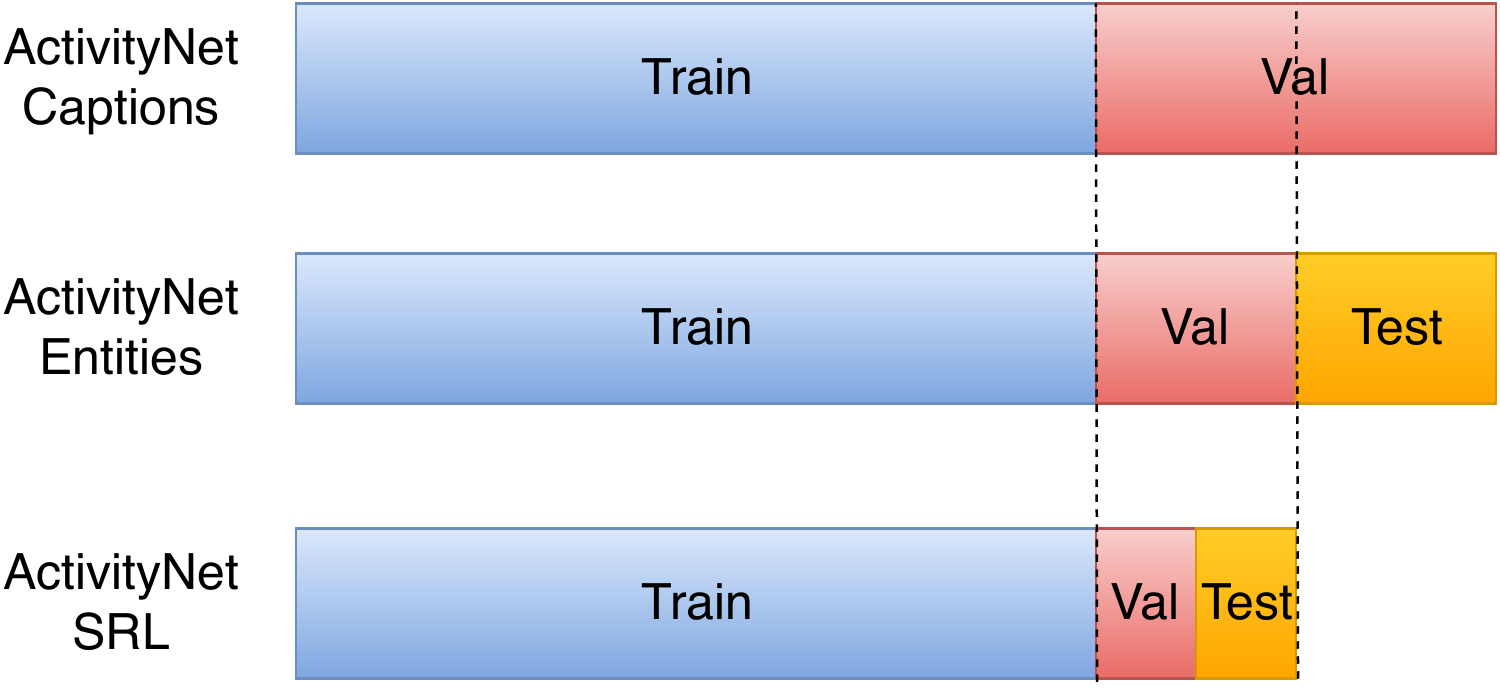}
    \caption{Train, val and test splits for AC, AE, ASRL.}
    \label{fig:sup_tvt_split}
\end{figure}

\textbf{Validation and Test:} Creating the validation and test splits is non-trivial.
Since the test split of AC is kept private, AE uses half of validation split of AC as its test split which is again kept private.
Thus, we divide the existing validation set into two to create the validation and test set for ASRL (see Figure \ref{fig:sup_tvt_split} for an illustration of deriving the splits, and Table \ref{tab:asrl_vid_stats} for the exact numbers).

Dividing the original validation set implies high miss-rate (\ie not enough examples to sample contrastive examples).
To address this, we allow contrastive sampling from the test set during validation and vice-versa during testing for more robust evaluation.

\subsection{Dataset Distribution}
\label{ss:sup_ds_dist}
\begin{figure}
    \centering
    \includegraphics[width=\linewidth]{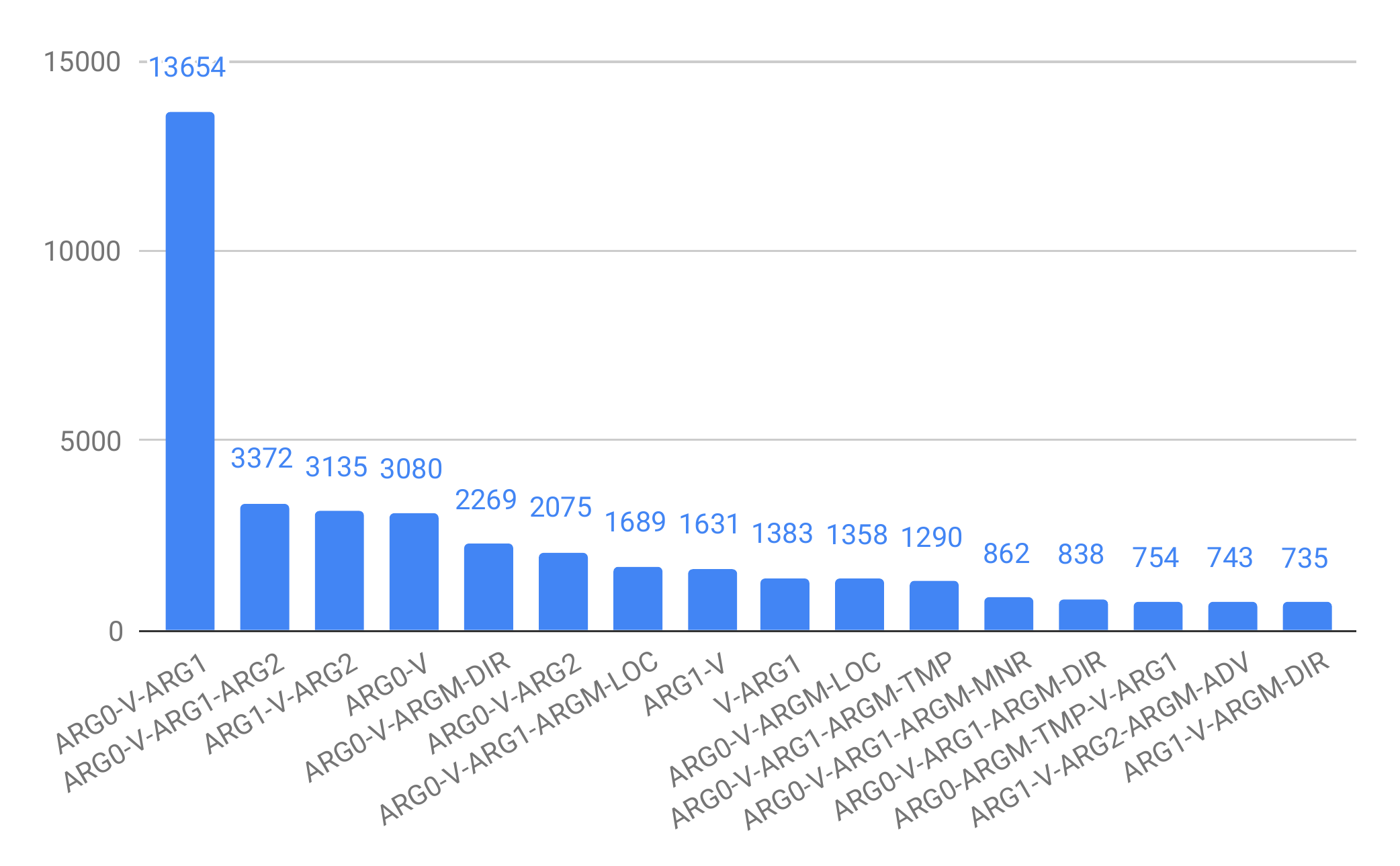}
    \caption{Frequently appearing SRL-Structures}
    \label{fig:srl_struct_freq}
\end{figure}

Figure \ref{fig:srl_struct_freq} highlights the distributions of the semantic-role-structures (\ie the order of the semantic role labels) found in the sentences.
It is interesting to note \ag{\dt{Arg0}}-\vb{\dt{Verb}}-\pat{\dt{Arg1}} far outnumbers all competing structures.
This also motivates the choice of considering $k{=}4$ videos at a time (if structure contains $3$ roles, we can sample $3$ more videos).

\begin{figure*}
    \centering
    \begin{minipage}{0.5\linewidth}
    \includegraphics[width=\linewidth]{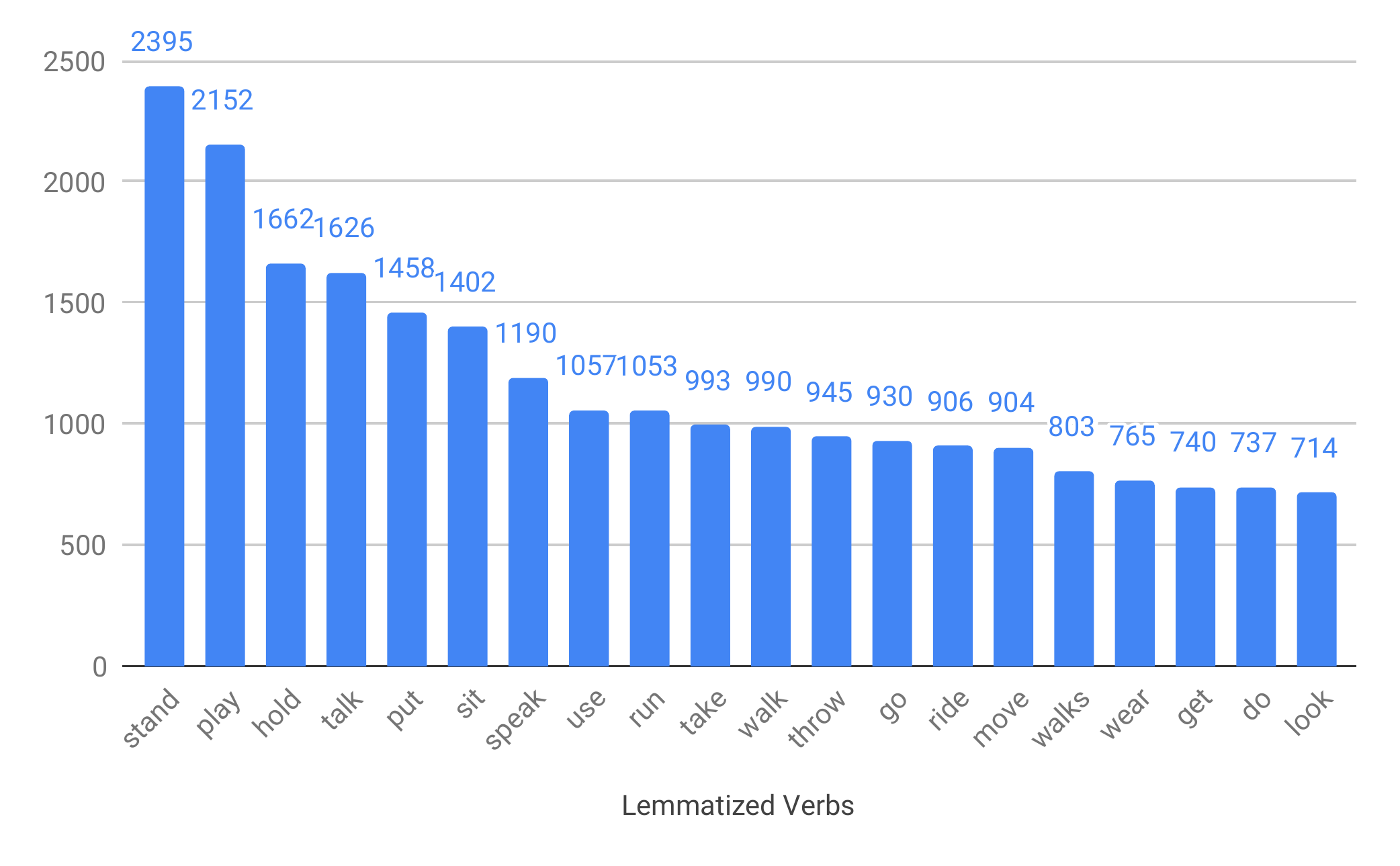}
    \caption{Top-20 Lemmatized \vb{\dt{Verb}}}
    \label{fig:lemma_verb}
    \end{minipage}%
    \begin{minipage}{0.5\linewidth}
    \includegraphics[width=\linewidth]{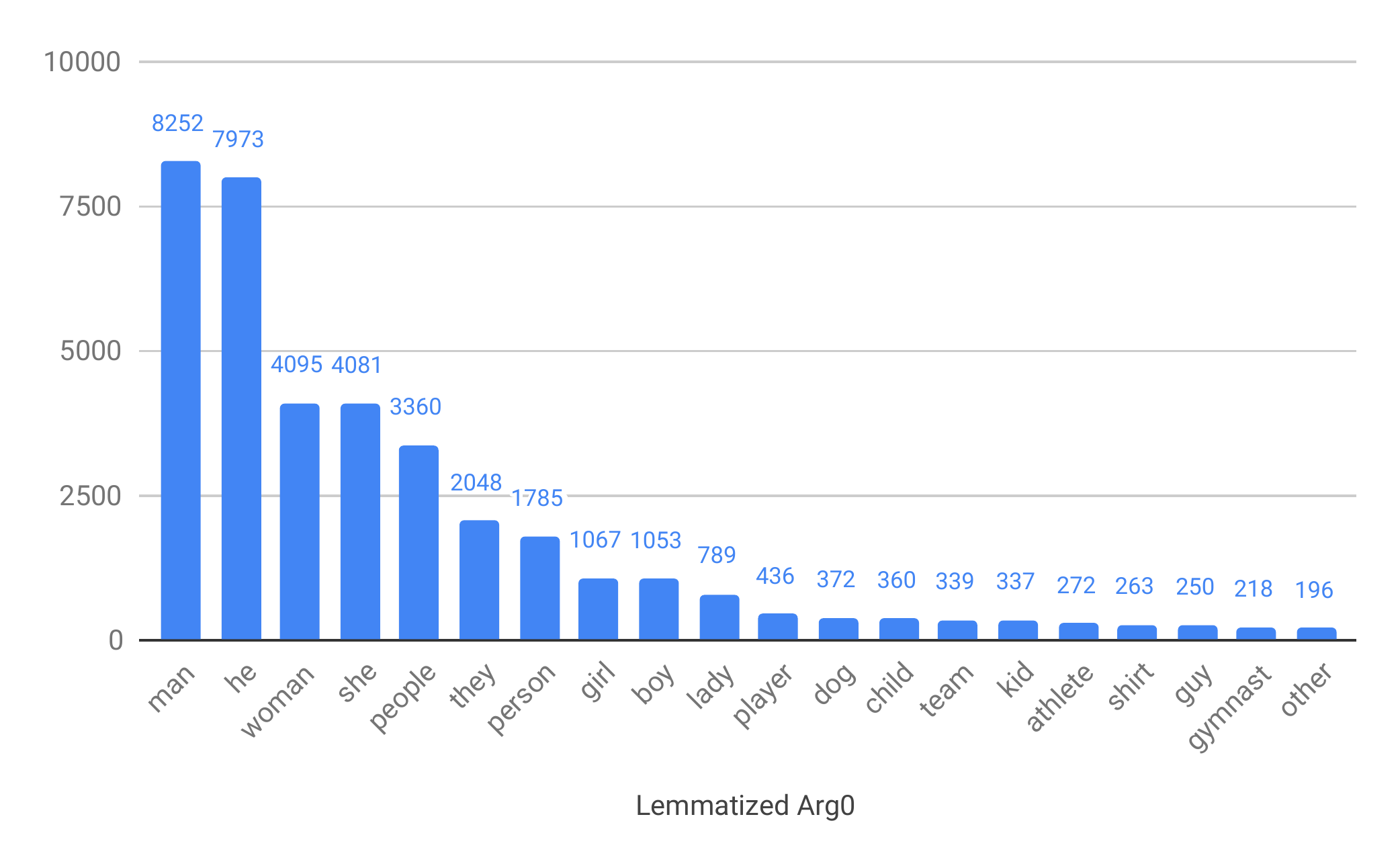}
    \caption{Top-20 Lemmatized \ag{\dt{Arg0}}}
    \label{fig:lemma_arg0}
    \end{minipage}
    \\
    \begin{minipage}{0.5\linewidth}
    \includegraphics[width=\linewidth]{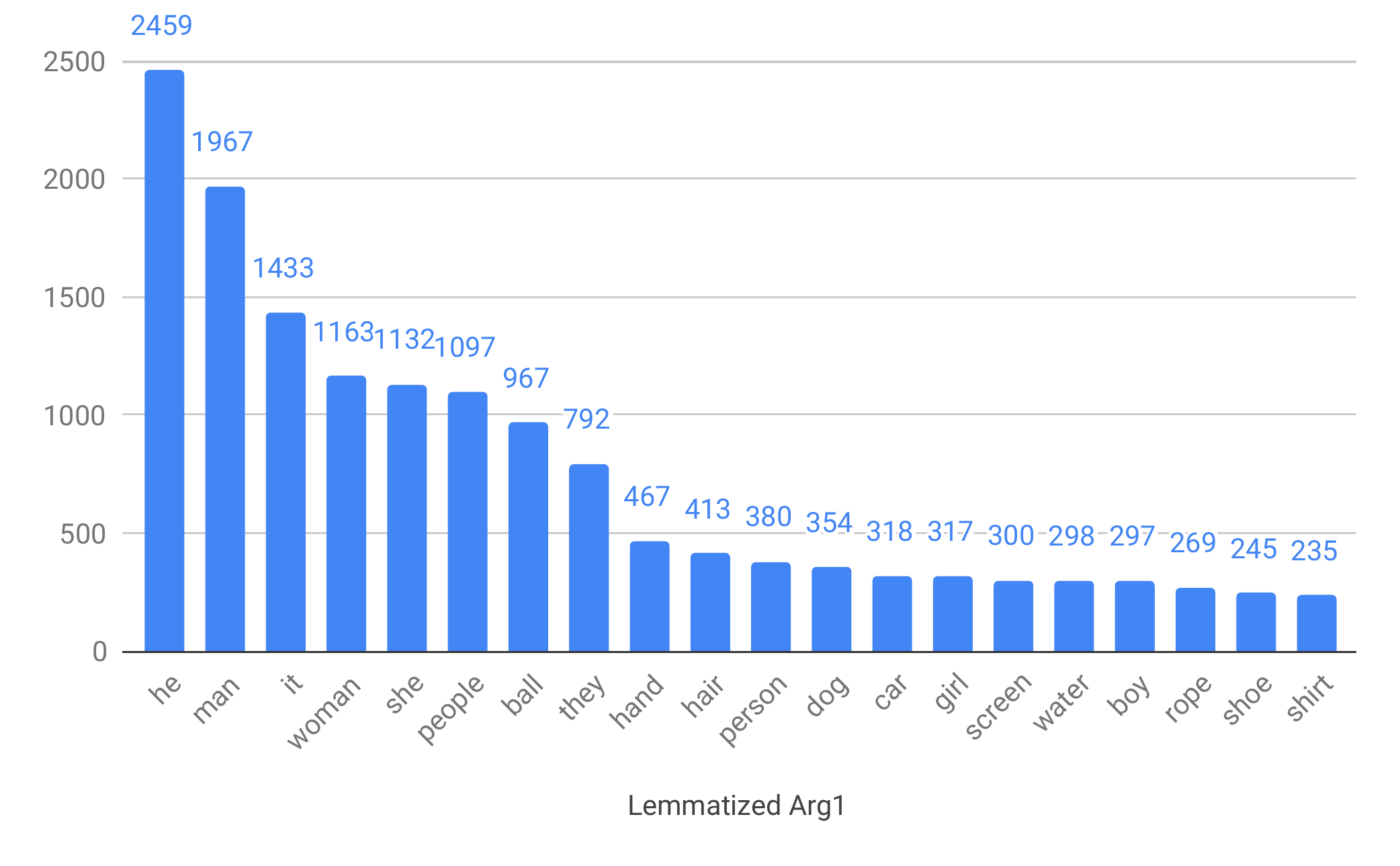}
    \caption{Top-20 Lemmatized \pat{\dt{Arg1}}}
    \label{fig:lemma_arg1}
    \end{minipage}%
    \begin{minipage}{0.5\linewidth}
    \includegraphics[width=\linewidth]{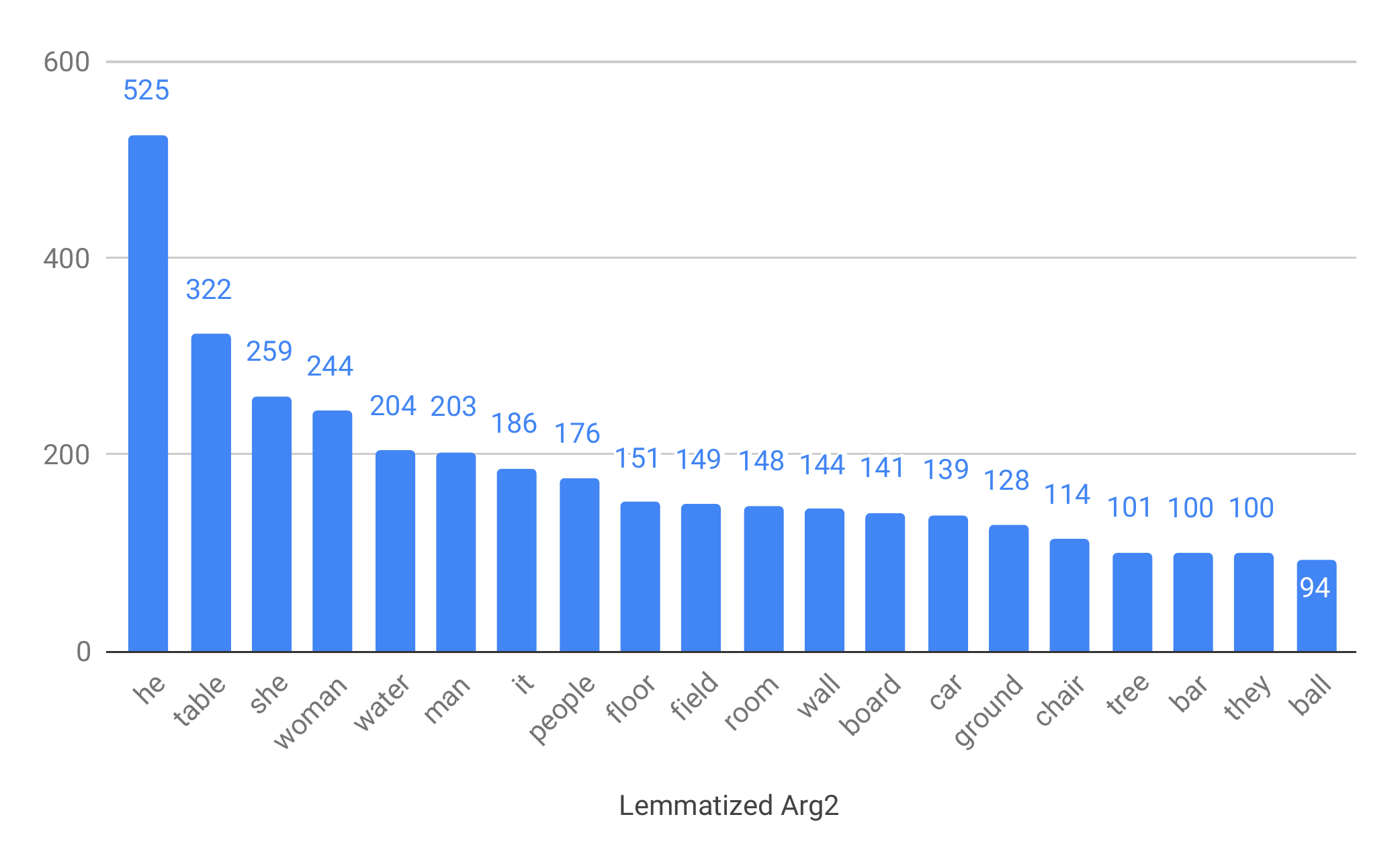}
    \caption{Top-20 Lemmatized \inst{\dt{Arg2}}}
    \label{fig:lemma_arg2}
    \end{minipage}

\end{figure*}
\begin{figure}
    \centering
    \includegraphics[width=\linewidth]{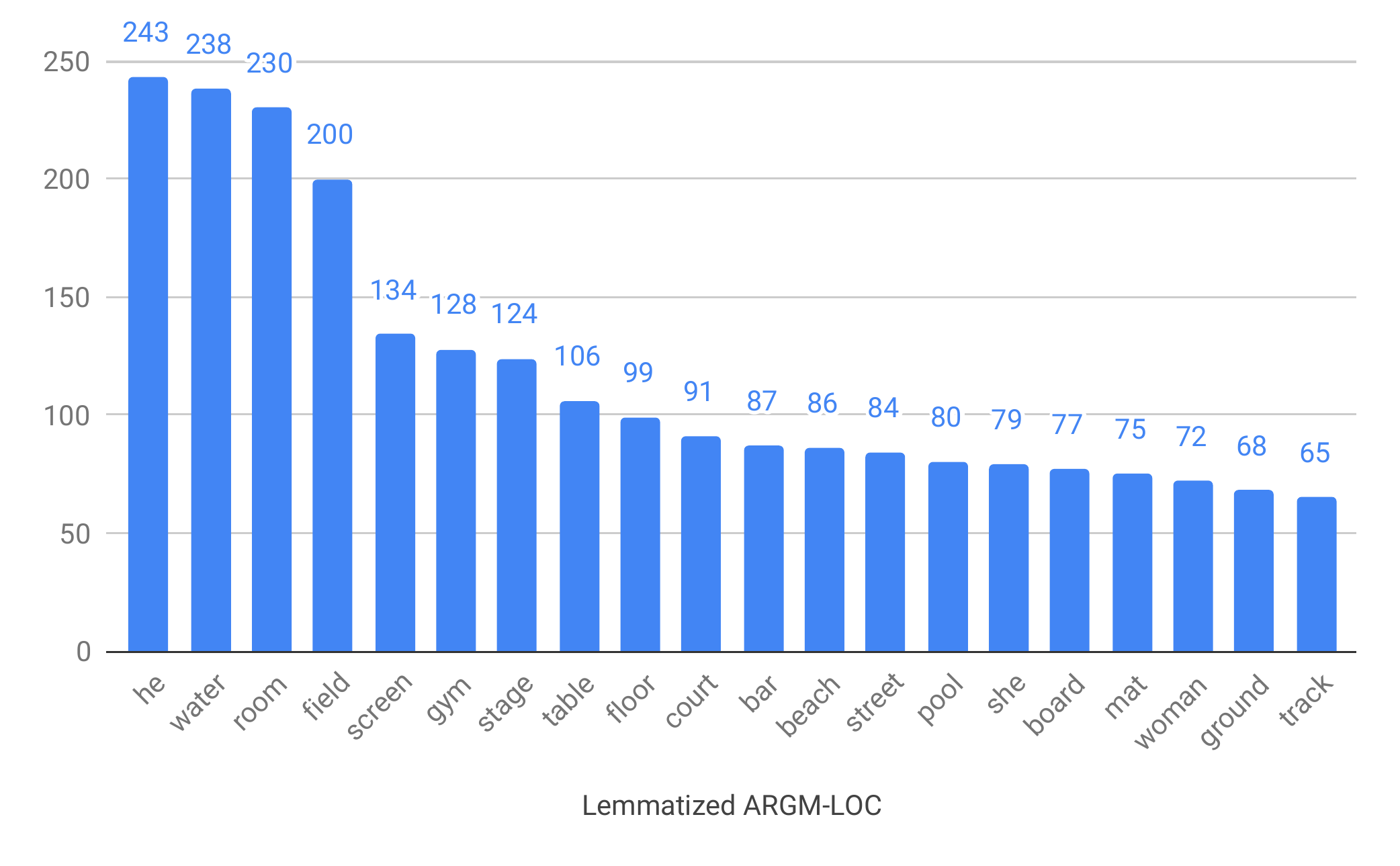}
    \caption{Top-20 Lemmatized \loc{\dt{ArgM-LOC}}}
    \label{fig:my_label}
\end{figure}

We look at the total number of lemmatized words in Table \ref{tab:sup_lemma_tot} and the most frequent (top-20) lemmatized words for each role with their frequencies: 
(i) \vb{\dt{Verb}} Figure \ref{fig:lemma_verb}
(ii) \ag{\dt{Arg0}} Figure \ref{fig:lemma_arg0}
(iii) \pat{\dt{Arg1}} Figure \ref{fig:lemma_arg1}
(iv) \inst{\dt{Arg2}} Figure \ref{fig:lemma_arg2}
.

The higher number of verbs (Table \ref{tab:sup_lemma_tot}) shows the diversity in the caption and their distribution is reasonably balanced (Figure \ref{fig:lemma_verb}) which we attribute to the curation process of ActivityNet \cite{caba2015activitynet}.
In comparison, \ag{\dt{Arg0}} is highly unbalanced as agents are mostly restricted to ``people''. 
We also observe that ``man'' appears much more often than ``woman''/``she''.
This indicates gender bias in video curation or video description.
Another interesting observation is that ``person'' class dominates in each of argument roles which suggest ``person-person'' interactions are more commonly described than ``person-object'' interactions.

\begin{table}[t]
\centering
\begin{tabular}{c|ccc}
\toprule
 & Training & Validation & Testing \\
\midrule
AC & 37421 & 17505 &  \\
AE & 37421 & 8774 & 8731 \\
ASRL & 31718 & 3891 & 3914 \\
\bottomrule
\end{tabular}
\vspace{1.5mm}

\caption{Number of Videos in train, validation, and test splits. Some instances are removed from training if they don't contain meaningful SRLs. Our test split is derived from AE validation set.
}
\label{tab:asrl_vid_stats}
\end{table}
\begin{table}[t]
\centering
\begin{tabular}{ccccc}
\toprule
V & Arg0 & Arg1 & Arg2 & ArgM-LOC \\
\midrule
338 & 93 & 281 & 114 & 59\\
\bottomrule
\end{tabular}
\vspace{1.5mm}
\caption{Total number of lemmatized words (with at least $20$ occurrence) in the train set of ASRL.}
\label{tab:sup_lemma_tot}
\end{table}

\subsection{Dataset Choice}
\label{ss:sup_ds_choice}
\textbf{Existing Datasets:} 
(As of Nov 2019) Other than ActivityNet, there are three video datasets which have visual and language annotations in frames namely EPIC-Kitchens \cite{Damen2018EPICKITCHENS}, TVQA+ \cite{Lei2019TVQASG} and Flintstones \cite{Gupta2018ImagineTS}.
We consider the pros and cons of each dataset.

EPIC-Kitchens contains ego-centric videos related to kitchen activity.
It provides object level annotations, with narrative descriptions.
While the annotations are rich, the narrative descriptions are too short in length (like ``open the fridge'' or ``cut the vegetable'') and the actors \ag{\dt{Arg0}} are not visible (ego-centric).

TVQA+ is a question-answering dataset subsampled from TVQA \cite{Lei2018TVQALC} with additional object annotations. 
While the videos are themselves rich in human activities, the questions are heavily dependent on the sub-titles which diminishes the role of actions.

Flintstones is a richly annotated dataset containing clips from the cartoon Flintstones.
The frames are 2-4 seconds long with 1-4 sentence description of the scene.
With the objects in cartoons easier to identify it also serves as a diagnostic dataset for video understanding.
However, the provided descriptions are less verb oriented and more image/scene-oriented due to shorter clips.

In contrast, ActivityNet contains longer videos, as a result the corresponding descriptions in ActivityNet Captions capture verbs over an extended period of time.
While the object annotations are richer in EPIC-Kitchens, TVQA+ and Flintstones, the richer verb-oriented language descriptions make it more suitable for our task.

\textbf{Using Natural Videos for evaluation:}
Our test data is generated ``synthetically'' by contrastive sampling followed by \spat and \temp strategies.
An alternative evaluation protocol would be to test on naturally occurring videos.
We discuss the challenges in obtaining such a dataset.

Recall that in our formulation of \tk a model needs to understand the relations among various objects prior to localizing them.
For instance, to evaluate if a model understands ``man petting a dog'' (example from Fig \ref{fig:ds4_evl_stg}a Q1,), we need contrastive examples Q2: ``X petting a dog'',Q3: ``man X a dog'',Q4: ``man petting X'' in the same video.
In the absence of any of these examples, it is hard to verify that the model indeed understands to query.
(\eg without Q3, ``man'' and ``dog'' could be localized without understanding ``petting'').
Creating such a test set from web sources is impractical at present because there is no large-scale densely annotated video dataset to isolate such contrastive videos.

A different (and quite expensive) method would be crowd-sourcing the video creation process by handing out detailed scripts to be enacted \cite{sigurdsson2016hollywood}. 
Here we would need to perform an additional ``domain adaptation'' step since we would be training and testing on different sources of videos (``YouTube'' $\rightarrow$ ``Crowd-Sourced'').
This makes it challenging to attribute the source of error \ie whether the reduced performance is due to poor generalization of object interactions or due to domain shift in the data.

In practice, \spat and \temp strategies when applied to contrastive videos from ActivityNet are effective proxies to obtaining naturally occurring contrastive examples from the web. This is validated by the drop from \svsq to \spat and \temp (Table \ref{tab:main_cmp}).

\vspace{\sectionReduceTop}
\section{Evaluation}
\vspace{\sectionReduceTop}
\label{sec:sup_eval}
We use the following evaluation metrics:
\begin{enumerate}
    \itemsep0em
    \item Accuracy: correct box is predicted for the given phrase in a sentence (a sentence has multiple phrases)
    \item Strict Accuracy: correct box is predicted for all the phrases in the sentence
    \item Consistency: predicted boxes (for all the phrases) belong to the same video, even if they are incorrect
    \item Video Accuracy: the predicted boxes are consistent, and the chosen video is also correct.
\end{enumerate}

Since there is only one video in \svsq, both consistency and video accuracy are not meaningful.
Similarly, we first choose a video in \sep, it is trivially consistent.

As mentioned earlier, the bounding box annotations in AE is sparse, the object has a bounding box in only one frame in the video where it is most clearly visible.
Since such sparse annotations complicate the computation of the above metrics,
we describe their computation for each case.

\subsection{Concatenation Strategies with Examples}
\begin{figure}
    \centering
    \includegraphics[width=\linewidth]{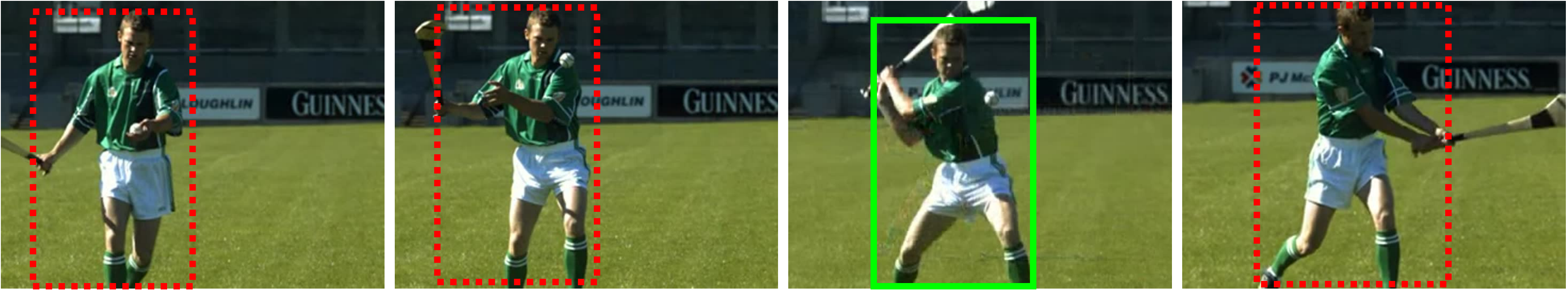}
    \caption{
    \svsq: Illustration of the ground-truth annotations for the ``man'' (green) obtained from AE. 
    The red boxes show equally correct boxes for ``man'' but are not annotated. As a result, we only consider the third frame to compute accuracy of the predicted box.
    }
    \label{fig:sup_svsq_gt}
    \vspace{1.5mm}
    \includegraphics[width=\linewidth]{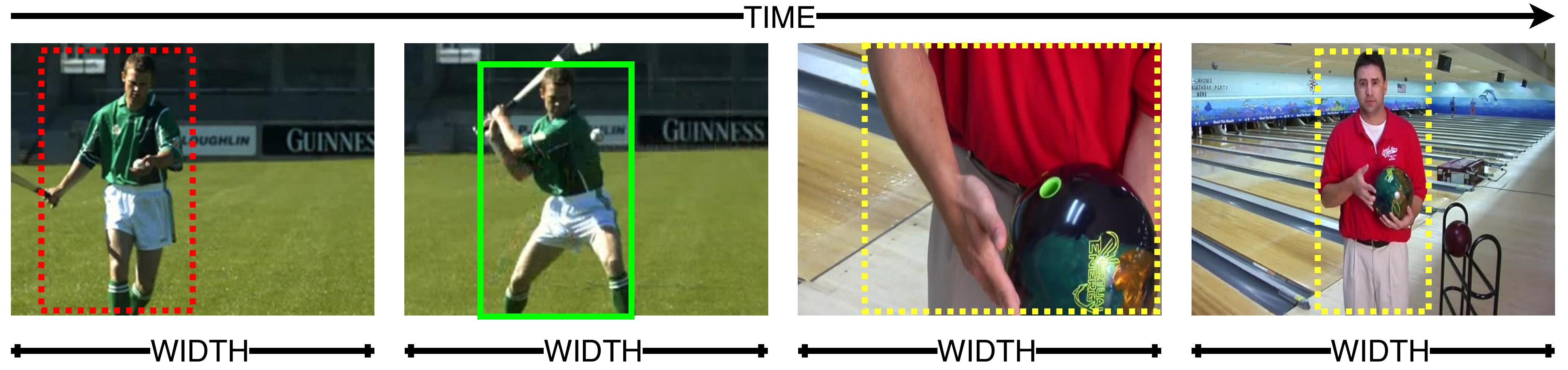}
    \caption{
    \temp: Two videos are concatenated along the time dimension (we show $2$ frames from each video) and with the description ``man throwing a ball'' and we are considering the object ``man''.
    If the predicted box is within the same video as ground-truth but the frame doesn't have any annotation (red box) we ignore it. 
    However, if the predicted box belongs to another video (yellow boxes), we say the prediction is incorrect. 
    }
    \label{fig:sup_temp_gt}
    \vspace{1.5mm}
    \includegraphics[width=\linewidth]{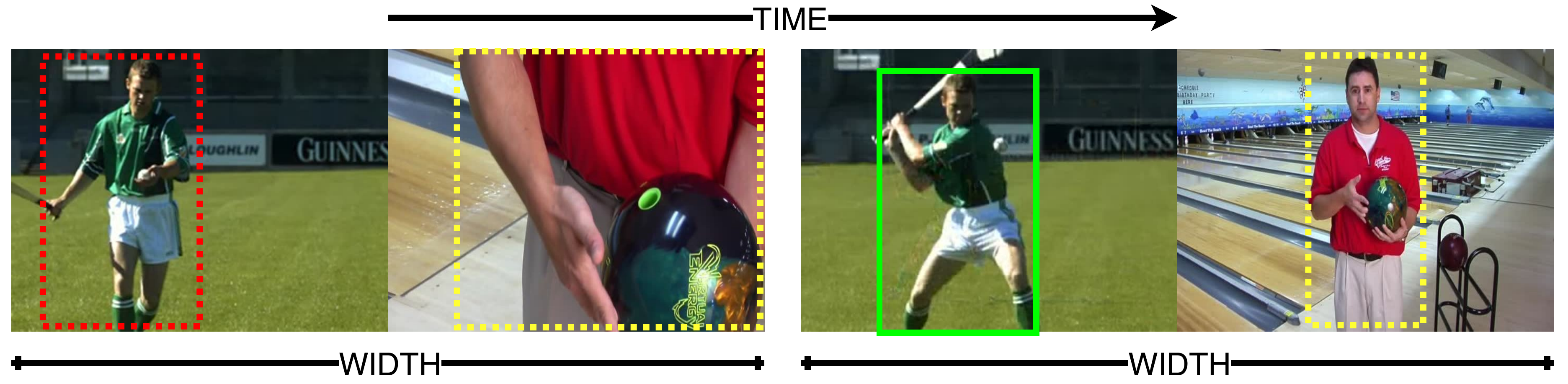}
    \caption{\spat: Similar to previous case, we have the same description of ``man throwing a ball'' and we consider the object ``man'' but the videos are concatenated along the width dimension (we show $2$ frames in the concatenated video).
    Again, if the predicted box lies in the same video as ground-truth (red box), we ignore it.
    If the predicted box is in another video (yellow boxes), the predictions are deemed incorrect.
    }
    \label{fig:sup_spat_gt}
\end{figure}
\underline{\textbf{\svsq}}:
We have a video with $F$ frames, however, for each object, the bounding boxes are available in exactly one frame.
Moreover, this annotated frame could be different for every object (the guideline provided in AE \cite{zhou2019grounded} is to annotate in the frame where it is most clearly visible).
As a result, we cannot be sure if the same object appears in a frame where it is not annotated.

To address this, we require the model to predict exactly one bounding box in every frame.
During evaluation, we consider only the annotated frame for a given object.
If in this annotated frame, there is a predicted bounding box with $IoU \ge 0.5$, we consider the object correctly predicted as illustrated in Figure \ref{fig:sup_svsq_gt}.
This gives us Accuracy for \svsq.
Strict Accuracy can be easily computed by considering all the phrases in the query \ie the predicted boxes for each phrase should have $IoU \ge 0.5$ with the ground-truth boxes.

\underline{\textbf{\sep}}:
We have $k$ videos and we choose one of these $k$ videos which gives us the Video Accuracy. 
If the chosen video is correct, we then apply scoring based on \svsq otherwise mark it incorrect.
Accuracy and Strict Accuracy computation is same as \svsq.

\underline{\textbf{\temp}}:
We have $k$ videos concatenated temporally.
In other words, we have $kF$ frames in total of which we know $(k-1)F$ frames don't contain the queried object.
Thus, if among the $(k-1)F$ frames not containing the queried object, if a predicted box has a score greater than a certain threshold, we mark it incorrect.
For the $F$ frames belonging to the queried video, we use the same method as for evaluating \svsq.
This is illustrated in Figure \ref{fig:sup_temp_gt}.

\underline{\textbf{\spat}}:
In \spat, we have $k$ videos concatenated along the width axis. 
That is, we have $F$ frames and each of width $kW \times H$ (here $W,H$ are the width and height of a single video).
In each of the $F$ frames, there should not be a predicted box outside the boundaries of the correct video with a score greater than some threshold and for the annotated frame the predicted box should have $IoU \ge 0.5$. 
This is illustrated in Figure \ref{fig:sup_spat_gt}.

For \temp and \spat strategies, Consistency is computed by how often the various objects are grounded in the same video.
Video Accuracy is derived from Consistency and is marked correct only when the correct video is considered.
Finally, Strict Accuracy measures when all the phrases in the query are correctly grounded.

\begin{table*}[t]
\centering
\resizebox{\textwidth}{!}{%
\begin{tabular}{cc|cc|ccc|cccc|cccc}
\tpr
\multirow{2}{*}{Model} & \multirow{2}{*}{Train} & \multicolumn{2}{c|}{\svsq} & \multicolumn{3}{c|}{\sep} & \multicolumn{4}{c|}{\temp}     & \multicolumn{4}{c}{\spat}    \\
                       &                        & \acc        & \sacc       & \acc   & \vacc  & \sacc  & \acc  & \vacc & \cons & \sacc & \acc & \vacc & \cons & \sacc \\
\midrule
\multirow{2}{*}{\bsi}                   & \gt                    & 46.31       & 24.83       & 20.55  & 47.49  & 9.92   & 8.06  & 2.68  & 25.35 & 2.68  & 4.64 & 2.47  & 34.17 & 1.31  \\
 & \phun                  & 55.22       & 32.7        & 26.29  & 46.9   & 15.4   & 9.71  & 3.59  & 22.97 & 3.49  & 7.39 & 4.02  & 37.15 & 2.72  \\
 \midrule
\multirow{2}{*}{\bsv}  & \gt                    & 43.37       & 22.64       & 22.67  & 49.6   & 11.67  & 9.35  & 3.37  & 28.47 & 3.29  & 5.1  & 2.66  & 33.6  & 1.74  \\
                       & \phun                  & 53.30       & 30.90       & 25.99  & 47.07  & 14.79  & 10.56 & 4.04  & 29.47 & 3.98  & 8.54 & 4.33  & 36.26 & 3.09  \\
\midrule
\multirow{2}{*}{\arch} & \gt                    & 46.25       & 24.61       & 24.05  & 51.07  & 12.51  & 9.72  & 3.41  & 26.34 & 3.35  & 6.21 & 3.40  & 39.81 & 2.18  \\
                       & \phun                  & 53.77       & 31.9        & 29.32  & 51.2   & 17.17  & 12.68 & 5.37  & 25.03 & 5.17  & 9.91 & 5.08  & 34.93 & 3.59 \\
\btr
\end{tabular}%
}

\vspace{1.5mm}
\caption{
Comparing models trained with \gt and \phun. All models are tested in \phun setting.
}
\label{tab:supp_main_cmp_full}
\end{table*}

\textbf{Selecting Threshold for Evaluation:}
As noted earlier, we pose the proposal prediction as a binary classification problem, if a proposal has a score higher than a threshold (hyper-parameter tuned on validation set), it is considered as a predicted box.
For evaluation, we consider only the boxes which have the highest score in each frame.
But in both \svsq and \sep cases there is no incentive to set a threshold (${>}0$), as the false positives cannot be identified in the same video.
On the other hand, in both \temp and \spat cases, false positives can be identified since we are sure boxes in a different video are negatives.

\vspace{\sectionReduceTop}
\section{Implementation Details}
\vspace{\sectionReduceTop}
\label{sec:sup_bsl}
\underline{\textbf{\bsi}} is an image grounding system that considers each frame separately.
It concatenates the language features to the visual features of each object which is then used to predict whether the given object is correct.
More specifically, given $\Tilde{q}_j$ (Eqn \ref{eq: srl embed}) and the visual features $\hat{v}_{i,j}$ we concatenate them to get the multi-modal features $m_{IG}$ where
$m_{IG}[l,i,j] = [\hat{v}_{i,j}||\Tilde{q}_l]$.
These are passed through a two-layered MLP classifier and trained using BCE Loss.
In essence, \bsi can be derived from \arch by removing the object transformer and the multi-modal transformer.

\underline{\textbf{\bsv}} is a video grounding system which builds upon \bsi.
Specifically, it has an object transformer to encode the language-independent relations among the objects.
More formally, given $\hat{v}_{i,j}$ we apply object transformer to get $\hat{v}_{i,j}^{sa}$.
The remaining steps are the same as \bsi.
We concatenate the language features $\Tilde{q}_j$ with the self-attended object features $\hat{v}_{i,j}^{sa}$ to get the multi-modal features $m_{VG}$ where $m_{VG}[l,i,j] = [\hat{v}_{i,j}^{sa}||\Tilde{q}_j]$.
After passing through a 2 layer MLP classifier, it is trained using BCE Loss.
In essence, \bsv can be derived from \arch by removing the multi-modal transformer altogether and the relative position encoding from object transformer.

\underline{\textbf{\arch}}:
Our models are implemented in Pytorch \cite{paszke2017automatic}.
\arch \spat using \gt takes nearly 25-30 mins per epoch (batch size $4$), compared to $3$ hours per epoch for \phun (batch size $2$).
All models are trained for 10 epochs (usually enough for convergence).
All experiments can be run on a single 2080Ti GPU.

\textbf{Language Feature Encoding}: We use a Bi-LSTM \cite{hochreiter1997long,Schuster1997BidirectionalRN} ( fairseq \cite{ott2019fairseq} implementation). 
The words are embedded in $\R^{512}$ and the Bi-LSTM contains $2$ layers with hidden size of $1024$, max sequence length of $20$, and $\mathcal{M}_q$ with input/output size of $256$.

\textbf{Visual Feature Encoding}: The object features are obtained from a FasterRCNN \cite{ren2015faster} with ResNext \cite{Xie2016AggregatedRT} pre-trained on Visual Genome \cite{krishna2017visual}. 
Each object feature is $2048$d vector.
The image level features ($2048$d) and optical flow ($1024$d) are extracted using resnet-200 \cite{he2016deep} and TVL1 \cite{zach2007duality} respectively and are encoded using temporal segment networks \cite{wang2016temporal}. 
They are concatenated to give segment features for each frame which are  $3072$d vector.
We project both object and segment features into $512$d vectors and then concatenate them to get $1024$d vector for each object.

\textbf{Object Transformer} uses $3$ heads and $1$ layer with each query, key, value of $1024$d (full feature dimension which is divided by number of heads for multi-headed attention).

\textbf{Multi-Modal Transformer} also uses $3$ heads and $1$ layer but the query, key, value are $1280$d vectors (additional $256$ due to concatenating with the language features).

\vspace{\sectionReduceTop}
\section{Additional Experiments}
\vspace{\sectionReduceTop}
\label{sec:sup_expts}

We perform two additional experiments:
(i) if the representations learned in \gt transfer to the more general case of \phun
(ii) the effect of adding more heads and layers to the object transformer (OTx) and multi-modal transformer (MTx).

\textbf{\gt models in \phun setting}:
In Table \ref{tab:supp_main_cmp_full} we compare 
the models \bsi, \bsv, and \arch trained in \gt and \phun and tested in \phun setting to calculate the transfer-ability of \gt setting.
While testing in \phun, for \temp and \spat, we set the threshold for models trained in \gt as $0.5$ which is higher than the threshold used when testing in \gt ($0.2$).
This is expected as a lower threshold would imply a higher chance of a false positive.

In general, the drop from \phun to \gt is significant (a $15{-}25\%$ drop) for almost all models suggesting training with just ground-truth boxes is insufficient.
Nonetheless, since the relative drops are same across models, \gt remains a valuable proxy for carrying out larger number of experiments.

\begin{table}[t]
\centering
\begin{tabular}{c|cccc}
\tpr
\spat & \acc & \vacc & \cons & \sacc \\
\midrule
\bsi            & 17.03 & 9.71  & 50.41 & 7.14 \\
+OTx (1L, 3H)   & 19.8  & 10.91 & 48.34 & 8.45  \\
+OTx (2L, 3H) & 20.8           & 11.38         & 49.45          & 9.17          \\
+OTx (2L, 6H) & \textbf{21.16} & \textbf{12.2} & 48.86          & \textbf{9.58} \\
+OTx (3L, 3H) & 20.68          & 11.34         & 48.66          & 9.19          \\
+OTx (3L, 6H) & 21.14          & 12.1          & \textbf{49.66} & 9.52  \\
\midrule
 \arch & 23.53 & 14.22 & 56.5  & 11.58 \\
+MTx (2L,3H) & 23.38 & 14.78 & 55.5  & 11.9  \\
+MTx (2L,6H) & 23.96 & 14.44 & 55.5  & 11.59 \\
+MTx (3L,3H) & 24.53 & 14.84 & 56.19 & 12.37 \\
+MTx (3L,6H) & 24.24 & 15.36 & 57.37 & 12.52 \\
\qquad+OTx(3L,6H) & \textbf{24.99} & \textbf{17.33} & \textbf{66.29} & \textbf{14.47} \\
\btr
\end{tabular}\vspace{1.5mm}
\caption{Ablative study layers and heads of Transformers.}
\label{tab:supp_tx_mabl}
\end{table}

\textbf{Transformer Ablation:}
In Table \ref{tab:supp_tx_mabl} we ablate the object transformer and the multi-modal transformer with number of layers and heads.
It is interesting to note adding more heads better than adding more layers for object transformer, while in the case of multi-modal transformer both number of heads and number of layers help.
Finally, we find that simply adding more layers and heads to the object transformer is insufficient, as a  multi-modal transformer with 1 layer and 3 heads performs significantly better than the object transformer with 3 layers and 6 heads.

\vspace{\sectionReduceTop}
\section{Visualization}
\vspace{\sectionReduceTop}
\label{sec:sup_vis}
\begin{figure*}
    \centering
    \includegraphics[width=\linewidth]{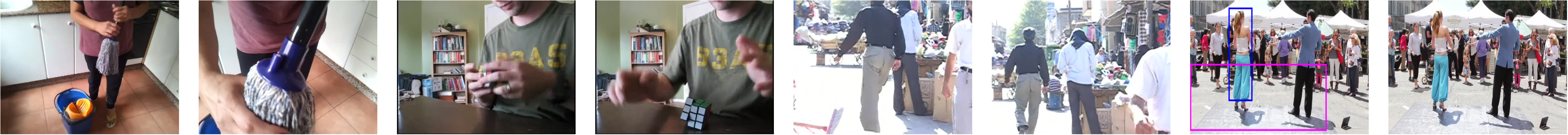}
    \\
    \begin{flushleft}
    \footnotesize{(a) Query: \pat{A woman} \vb{standing} on a \inst{sidewalk}.
    From left to right, other videos are: 
    (1): A woman standing in kitchen
    (2): A man solving a puzzle
    (3): Men standing on sidewalk.
    Our model disambiguates the two ``sidewalks'', as well as the ``woman'' and localizes them in the same video.
    Here (2) is a randomly sampled (``woman'', ``sidewalk'' only have ``stand'' relation).}
    \end{flushleft}
    \includegraphics[width=\linewidth]{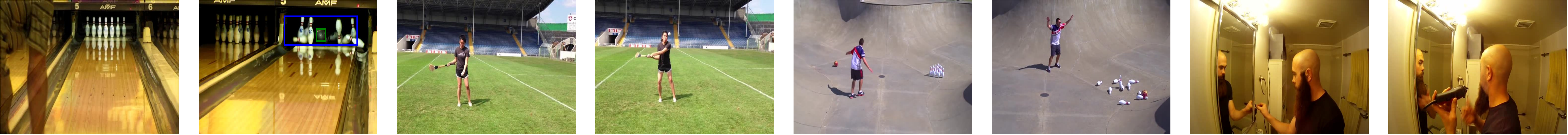}
    \\
    \begin{flushleft}
    \footnotesize{
    (b) Query: \ag{The ball} \vb{hits} the \pat{pins} creating a strike. 
    From left to right, other videos are:
    (2): The girl with the ball hits it
    (3): A bowling ball hits the pins.
    (4): He uses razor to trim.
    While our model correctly chooses the correct frame, we note (3) is very close to (1) in terms of description.
    Here, our sampling method fails by providing ``too'' similar videos.
    }
    \end{flushleft}
    \caption{\arch predictions \temp strategy in \gt setting.
    We show two frames from each video, but the model looks at $F{=}40$ frames.
    }
    \label{fig:sup_temp_pos}
\end{figure*}

\begin{figure}[t]
    \centering
    \includegraphics[width=\linewidth]{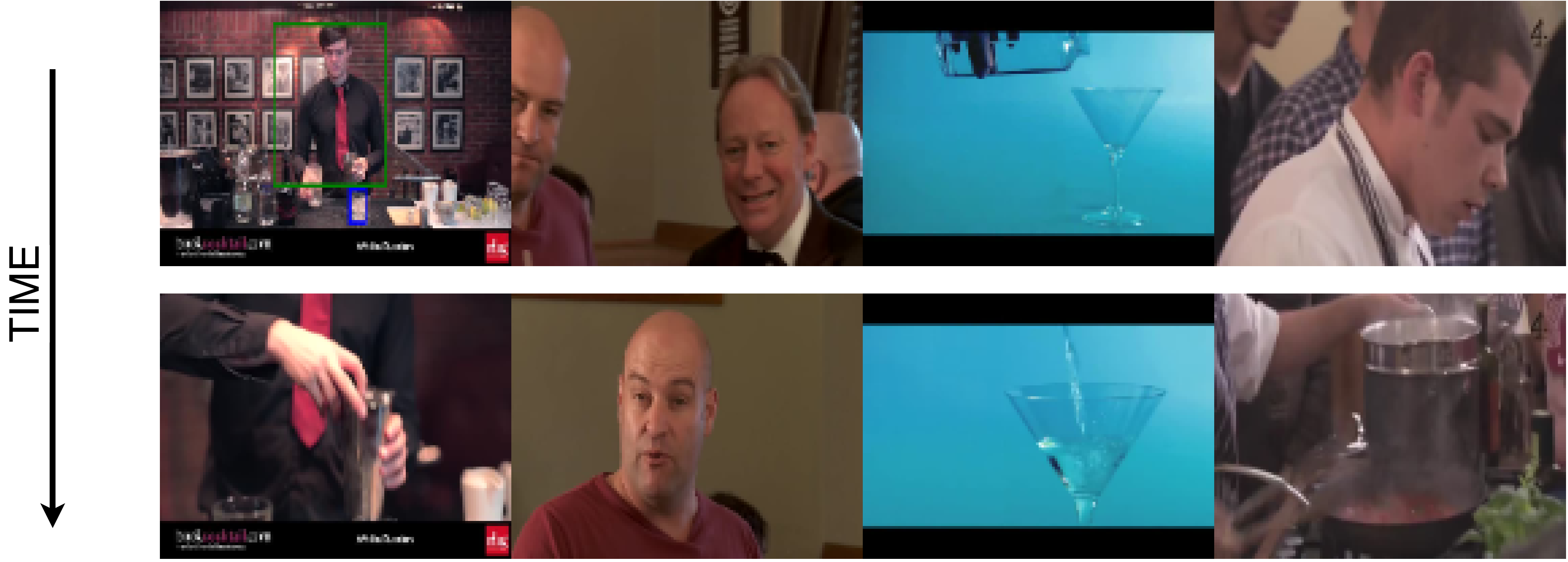}
    \\
    \begin{flushleft}
    \footnotesize{
    (a) Query: \ag{He} \vb{pours} the \pat{mixed drink} [\dt{Arg3}: to the cup].
    Left-to-right other videos are:
    (2): Two men drinking an energy drink.
    (3): A drink poured into martini glass.
    (4): A young man pours oil into the pan.
    Our model finds the ``man'' and the ``mixed drink'' correctly but fails to localize the ``cup'' due to small number of queries containing Arg3.
    }
    \end{flushleft}
    \includegraphics[width=\linewidth]{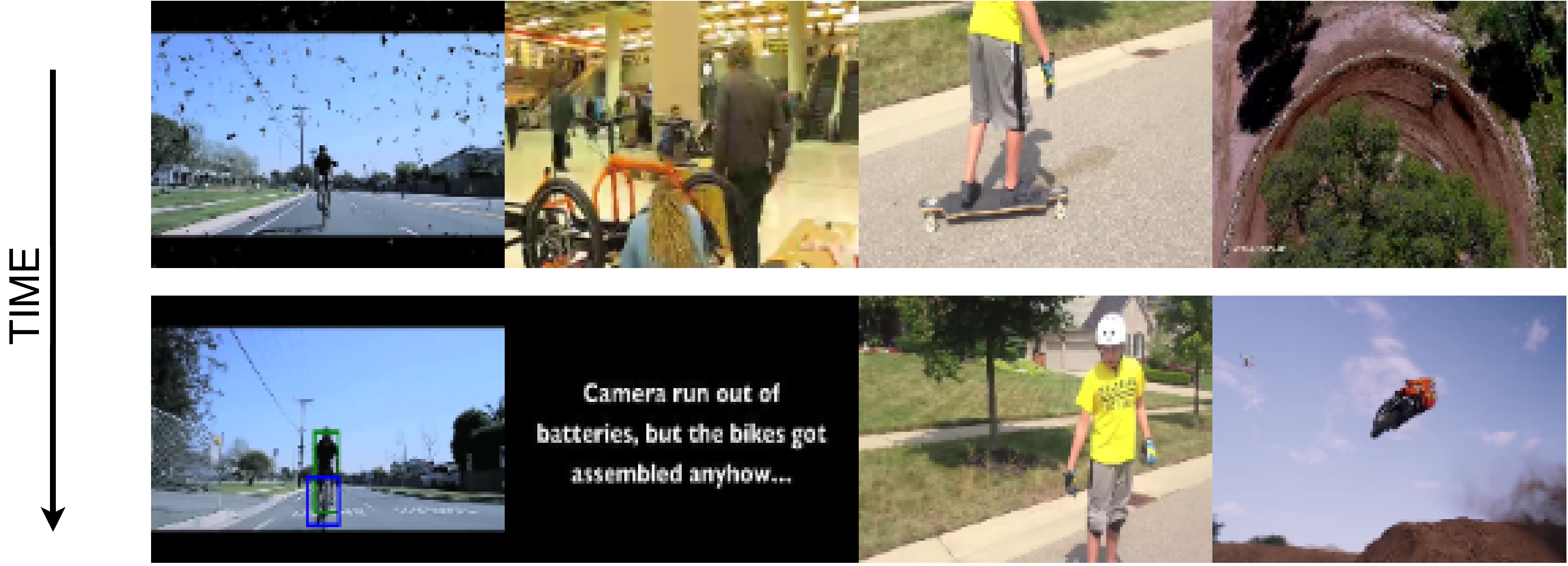}
    \\
    \begin{flushleft}
    \footnotesize{
    (b) Query: \ag{The man} \vb{riding} \pat{the bike}.
    Left-to-right other videos are:
    (2): Men put the other bike down.
    (3): A boy rides his skateboard.
    (4): We see the boy riding his dirtbike.
    Here, our model correctly distinguishes among the bikes, and who is riding what.
    }
    \end{flushleft}
    \caption{\arch predictions \spat strategy in \gt setting.
    We show two frames from each video, and each frame contains $4$ videos concatenated together.}
    \label{fig:spat_pos1_vis}
\end{figure}

\begin{figure}
    \centering
        \includegraphics[width=\linewidth]{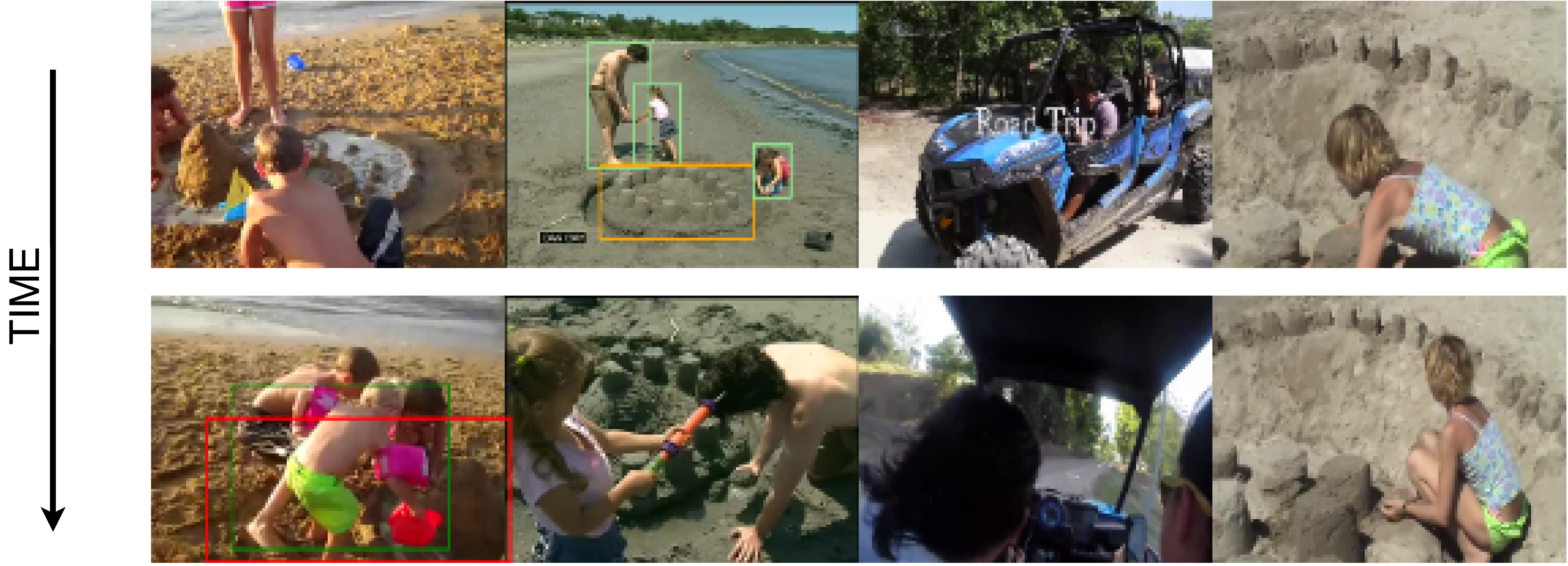}
    \\
    \begin{flushleft}
    \footnotesize{
    Query: \ag{A man and two kids} \vb{building} \pat{a sand castle}. 
    Left-to-right other videos: 
    (1): A group of kids trying to build a sand castle.
    (3): They are driving through the sand.
    (4): She is building a castle.
    In the first frame, the ground-truths are marked in light-green and orange and in the second frame is our model's incorrect prediction.
    It is unable to distinguish based on ``man'' due to influence of ``kids'' in the agent.
    }
    \end{flushleft}
    \caption{Incorrect prediction of \arch for \spat strategy}
    \label{fig:spat_neg1_vis} 
\end{figure}

In general, contrastive examples differ in exactly one part of the phrase. 
However, we observed that some contrastive examples were very difficult to distinguish. 
We identify two reasons: 
(i) Considering only one verb in the query becomes restrictive. 
For instance, in Figure \ref{fig:sup_temp_pos}-(b) video (3), the complete description has ``the bowling ball that goes around the ring and then hits the pins'' and the initial part of it going around the ring is lost.
(ii) Language ambiguity of the form ``person playing guitar'' vs ``person practicing guitar'', while ``playing'' and ``practicing'' have distinct meanings, in some situations they can be used interchangeably.

We now visualize a few examples for \temp and \spat in Figure \ref{fig:sup_temp_pos}, \ref{fig:spat_pos1_vis}, \ref{fig:spat_neg1_vis}.
All visualizations are obtained using \arch trained in \gt setting.
For each case, we show $2$ frames from each video and color-code the arguments in the given query (Arg0 is Green, Verb is Red, Arg1 is Blue, Arg2 is Magenta) 
Remaining arguments are mentioned in the query (like in Figure \ref{fig:spat_pos1_vis} (a)).

For \temp, since objects are not being considered independent of each other, the model doesn't ground objects which are present in the query but not related. 
For instance in Figure \ref{fig:sup_temp_pos}-(a), even though ``woman'' and ``sidewalk'' are separately present in two other videos, these are given very low score. 
Similarly, in Figure \ref{fig:sup_temp_pos}-(b), ``ball'' in (2) is not grounded which is not related to the query.
These suggest \arch is able to exploit the cues in the language query to ground the objects and their relations in the visual domain.

For \spat, in Figure \ref{fig:spat_pos1_vis}-(a) our model finds the correct video.
It is able to differentiate among someone pouring drink into a glass (2), someone pouring oil (3), or someone drinking the drink (1).
However, it is unable to find the ``cup'' which we attribute to the smaller number of examples containing \dt{Arg3} which is limited to verbs like ``pour''.
In Figure \ref{fig:spat_pos1_vis}-(b) our model correctly finds both ``man'' and the ``bike'' that he is riding and distinguishes between ``ride'' and ``put'', ``bike'' and ``skateboard'' (3).

Finally, in Figure \ref{fig:spat_neg1_vis}, we find the language ambiguity of ``trying to build'' and ``building'' which are synonymously used.
While our model is able to distinguish (4) by its agent ``she'' compared to ``man and two kids'',
it is unable to make the distinction between ``a man and two kids'' and ``a group of kids'' (1).
We attribute this to the use of a single embedding for each role (here \ag{Arg0}) and not differentiating among the various objects in that role.

{\small
\bibliographystyle{ieee_fullname}
\bibliography{ref,lang_ref,new_ref,vid_ref}
}

\end{document}